\documentclass[pdflatex,sn-mathphys-num]{sn-jnl}

\usepackage{graphicx}%
\usepackage{multirow}%
\usepackage{amsmath,amssymb,amsfonts}%
\usepackage{amsthm}%
\usepackage{mathrsfs}%
\usepackage[title]{appendix}%
\usepackage{xcolor}%
\usepackage{textcomp}%
\usepackage{manyfoot}%
\usepackage{booktabs}%
\usepackage{algorithm}%
\usepackage{algorithmicx}%
\usepackage{algpseudocode}%
\usepackage{listings}%
\usepackage{csquotes}
\usepackage{soul}
\usepackage{xcolor}
\soulregister{\cite}{7}
\soulregister{\ref}{7}
\soulregister{\pageref}{7}

\begin{document}

\title[Article Title]{Rehabilitation Exercise Quality Assessment through Supervised Contrastive Learning with Hard and Soft Negatives}


\author[1]{\fnm{Mark} \sur{Karlov}}\email{mark.karlov@mail.utoronto.ca}
\equalcont{These authors contributed equally to this work.}

\author*[2]{\fnm{Ali} \sur{Abedi}}\email{ali.abedi@uhn.ca}
\equalcont{These authors contributed equally to this work.}

\author[2]{\fnm{Shehroz S.} \sur{Khan}}\email{shehroz.khan@uhn.ca}

\affil*[1]{\orgdiv{Department of Electrical and Computer Engineering}, \orgname{University of Toronto}, \orgaddress{\street{10 King's College Road}, \city{Toronto}, \postcode{M5S 3G4}, \state{Ontario}, \country{Canada}}}

\affil[2]{\orgdiv{KITE Research Institute}, \orgname{University Health Network}, \orgaddress{\street{550 University Avenue}, \city{Toronto}, \postcode{M5G 2A2}, \state{Ontario}, \country{Canada}}}



\abstract{Exercise-based rehabilitation programs have proven to be effective in enhancing the quality of life and reducing mortality and rehospitalization rates. AI-driven virtual rehabilitation, which allows patients to independently complete exercises at home, utilizes AI algorithms to analyze exercise data, providing feedback to patients and updating clinicians on their progress. These programs commonly prescribe a variety of exercise types, leading to a distinct challenge in rehabilitation exercise assessment datasets: while abundant in overall training samples, these datasets often have a limited number of samples for each individual exercise type. This disparity hampers the ability of existing approaches to train generalizable models with such a small sample size per exercise type. Addressing this issue, this paper introduces a novel supervised contrastive learning framework with hard and soft negative samples that effectively utilizes the entire dataset to train a single model applicable to all exercise types. This model, with a Spatial-Temporal Graph Convolutional Network (ST-GCN) architecture, demonstrated enhanced generalizability across exercises and a decrease in overall complexity. Through extensive experiments on three publicly available rehabilitation exercise assessment datasets, UI-PRMD, IRDS, and KIMORE, our method has proven to surpass existing methods, setting a new benchmark in rehabilitation exercise quality assessment.}




\maketitle

\section{Introduction}\label{sec1}
Patients undergoing treatments related to cardiac, stroke, and other injuries are often referred to rehabilitation programs for swift recovery. These programs aim to enhance the quality of life of these patients, improving their ability to live independently and reducing their risk of re-hospitalization and death \cite{WHORehabilitation}. Typically, these programs emphasize prescribed exercises to restore mobility, muscle mass, and overall bodily strength \cite{dibben2023exercise,frazzitta2013beneficial,liao2020review}. Traditionally provided in a clinical setting, these programs often face long wait times, staff shortages, and logistical challenges, such as transportation and scheduling \cite{shanmugasegaram2012psychometric,shirozhan2022barriers,combes2018hospital}. Virtual and home-based rehabilitation \cite{ferreira2023usage,combes2018hospital,krasovsky2020will} offers a flexible alternative that overcomes these challenges and delivers benefits akin to in-clinic sessions \cite{seron2021effectiveness,boukhennoufa2022wearable}. Leveraging data from virtual rehabilitation sessions, Artificial Intelligence (AI) can assess the quality of exercises, the patient's recovery progression, and their risks of dropping out of the programs \cite{abedi2024artificial}. These approaches may utilize a variety of sensors, including wearable sensors or cameras, to monitor patients' movements. AI algorithms analyze this data during exercises \cite{ferreira2023usage,sangani2020real,sardari2023artificial}. This analysis provides valuable feedback to patients on exercise quality and completion while enabling clinicians to track progress and personalize interventions effectively \cite{sangani2020real,fernandez2018virtualgym,sardari2023artificial,abedi2024artificial}.

Rehabilitation exercise quality assessment is grounded in objective measures. These include adherence to prescribed sets and repetitions of exercises \cite{10229850,abedi2023cross}, consistency in exercise execution, ensuring proper technique and quality of movement, and maintaining correct posture of different body parts \cite{kimore,liao2020review,li2024finerehab,capecci2018instrumental}. The development of AI models for rehabilitation exercise quality assessment relies on annotated datasets \cite{yan2018spatial,yao2023contrastive,zheng2023skeleton,deb2022graph,liao2020review}. Experts in the field, such as rehabilitation clinicians and physiotherapists, observe patients as they complete exercises and annotate them \cite{li2024finerehab,kimore}. The exercise data and their corresponding annotations are then used for the development of AI models. In some datasets, such as the KInematic assessment of MOvement and clinical scores for remote monitoring of physical REhabilitation (KIMORE) \cite{kimore}, clinically validated tools, such as the Exercise Accuracy Assessment Questionnaire (EAAQ) \cite{capecci2018instrumental}, were used for annotation resulting in real-valued numbers representing exercise quality scores. However, other datasets , such as the University of Idaho-Physical Rehabilitation Movement Data (UI-PRMD) \cite{prmd} and IntelliRehabDS (IRDS) \cite{irds}, sufficed with annotating rehabilitation exercises as correct or incorrect binary values. This inconsistency \cite{khan2022inconsistencies} hinders the development of AI models applicable across datasets. A model trained on IRDS for a binary classification problem is not directly applicable to KIMORE, which requires solving a regression problem.

Current methods in automated rehabilitation exercise quality assessment mainly utilize three types of data: acceleration data from inertial wearable sensors, video data from RGB or depth cameras, and body joint data obtained either through sensors such as Kinect or extracted from RGB videos using computer vision techniques \cite{pavllo20193d,lugaresi2019mediapipe,abedi2023cross} \cite{liao2020review,sardari2023artificial}. Past studies in general human activity analysis have underscored the significance of analyzing body joints \cite{yan2018spatial,liao2020review}. In rehabilitation exercise quality assessment, this body joint analysis approach mirrors the methods clinicians use to evaluate exercise quality and technique \cite{kimore}. Body joints are less affected by changes in lighting and background, making them a stable data source for analysis. This paper focuses on assessing exercises based on sequences of body joints through space and time using Spatial-Temporal Graph Convolutional Networks (ST-GCNs) \cite{yan2018spatial,yao2023contrastive,zheng2023skeleton,deb2022graph}.

Patients in rehabilitation programs receive group and individual level exercises that are tailored to their specific needs, stage of their rehabilitation program, age, sex, and comorbidity \cite{frazzitta2013beneficial}. Examples of rehabilitation exercise types include standing shoulder abduction, right elbow flexion, and deep squats \cite{prmd,irds,kimore}. A primary challenge in current methods of rehabilitation exercise quality assessment \cite{yan2018spatial,yao2023contrastive,zheng2023skeleton,deb2022graph} lies in their dependency on distinct models for each type of rehabilitation exercise. Training individual models for each exercise type is problematic. Existing datasets, such as UI-PRMD \cite{prmd}, often have a high total number of samples that are spread sparsely across various exercise types, resulting in insufficient samples for each specific exercise type. This poses a challenge for training exercise-type-specific deep neural networks that require large amounts of data \cite{liao2020review,sardari2023artificial,khanghah2023novel}.

This paper aims to address the above-mentioned issue by proposing a unified model that leverages all samples across different exercise types in a dataset, rather than separate models limited to their respective exercise samples. This approach considers that each exercise type falls within a specific range of acceptable latent spaces for correct assessment, with deviations indicating incorrectness. This principle is uniformly applicable across all exercise types using a single exercise assessment model. To this end, this paper makes two key contributions: (i) introducing a novel supervised contrastive learning method \cite{supcon} equipped with hard \cite{robinson2021contrastive} and soft negatives specifically designed for rehabilitation exercise quality assessment, where a single ST-GCN model abstracts the general assessment process in a dataset, and (ii) demonstrating quantitative superiority over previous methods, enhancing the state-of-the-art in rehabilitation exercise quality assessment on three public datasets, UI-PRMD \cite{prmd}, IRDS \cite{irds}, and KIMORE \cite{kimore}. In an attempt to develop an exercise quality assessment model applicable across datasets, a model was trained on IRDS \cite{irds} through the proposed contrastive learning approach. This model then underwent transfer learning to make inferences on KIMORE \cite{kimore} as a regression problem, showing advancements compared to previous works.

The organization of this paper is as follows. Section \ref{sec:related_work} provides an overview of related literature. This is followed by Section \ref{sec:methodology}, where the methodology we propose is detailed. Subsequently, Section \ref{sec:experiments} outlines the experimental settings and discusses the results obtained using the proposed method. Finally, Section \ref{sec:conclusion} concludes the paper and suggests avenues for future research.

\section{Related Work}
\label{sec:related_work}
This section reviews existing approaches for rehabilitation exercise quality assessment, including both general deep learning techniques (non-ST-GCN) and ST-GCN-based methods specifically designed for assessing rehabilitation exercise quality based on body joints \cite{liao2020review,sardari2023artificial}. With regard to the focus and contributions of the method introduced in this paper, this section explores strategies previously employed to address the specific settings of existing datasets for rehabilitation exercise quality assessment. These settings typically feature a variety of exercise types with a limited number of samples for each type \cite{liao2020review,sardari2023artificial,prmd,irds,kimore}.

\subsection{General Deep Learning Methods}
\label{sec:related_work_general}
Liao et al. \cite{liao2020deep} developed one of the first deep neural networks for rehabilitation exercise quality assessment. Their initial step involved reducing the dimensionality of the input data, namely, the number of body joints, using various methods, including maximum variance, principal component analysis \cite{bashir2005hmm}, and Long Short-Term Memory (LSTM) autoencoders. The deep-learning model for exercise assessment consists of parallel temporal pyramid sub-networks and cascades of LSTM layers. The temporal pyramid sub-networks apply 1D convolutions to sequences of body joints (or their reduced-dimensionality versions) at different time resolutions and then concatenate the convolution results. The multi-layer LSTM network analyzes the output from the temporal pyramid sub-networks and outputs the quality of rehabilitation exercises. The first drawback of this method is its separation of the dimensionality reduction module from the exercise quality assessment module, rather than developing them jointly. Joint learning could lead to an understanding of which body joints are more effective at differentiating between correct and incorrect exercises \cite{lin2023actionlet}. The second drawback is the overlooking of the spatial relationship among body joints, treating the sequence of body joints as a multivariate time series. The third drawback is the necessity for exercise-type-specific models. For instance, for the UI-PRMD dataset \cite{prmd}, ten separate exercise-type-specific models were developed. These exercise-type-specific models could not leverage the samples of all exercise types in the entire dataset and were trained on data from single exercise types, which are limited to a certain number.

In the method proposed by Abedi et al. \cite{abedi2023cross}, the first step involved extracting body joints from rehabilitation exercise videos using MediaPipe \cite{lugaresi2019mediapipe}. This was followed by the extraction of exercise-type-specific features \cite{guo2021exercise} from the sequences of body joints, which were then input into exercise-type-specific LSTM models for rehabilitation exercise quality assessment. To increase the training dataset's size and enhance the generalizability of the models, cross-modal video-to-body-joints augmentation techniques were employed on the KIMORE dataset \cite{kimore}. Techniques for visual augmentation were applied to the video data, and the body joints extracted from the augmented videos were utilized in training the models for specific exercise types. The experimental findings on the KIMORE dataset \cite{kimore} demonstrated a notable improvement in rehabilitation exercise quality assessment following the cross-modal augmentation. Building upon the pipeline developed by Abedi et al. \cite{abedi2023cross}, Karagoz et al. \cite{karagoz2023supervised} utilized supervised contrastive learning to train exercise-type-specific LSTM models for rehabilitation exercise quality assessment. The original supervised contrastive learning \cite{supcon} was modified \cite{zha2024rank} to handle the imbalanced distribution of samples of specific exercise types in the KIMORE dataset \cite{kimore}. Despite not surpassing previous methods in performance, the use of supervised contrastive loss was noted to outperform the L1 loss on the KIMORE dataset \cite{kimore}. In \cite{abedi2023cross} and \cite{karagoz2023supervised}, the pipeline comprised extracting handcrafted features from the body joints in video frames, designed specifically for different types of exercise. Despite the selection of handcrafted features being based on the angles between body joints for exercises, the method did not fully account for the spatial relationships among body joints. Additionally, the method limited the training of deep learning models to only include samples from specific types of exercises in the dataset, instead of enabling training across exercise types.

\subsection{ST-GCN-based Methods}
\label{sec:related_work_stgcn}
ST-GCNs \cite{yan2018spatial} have been widely used for skeleton-based action analysis, including action recognition and classification \cite{yan2018spatial,lin2023actionlet}. Deb et al. \cite{deb2022graph} explored ST-GCN's role in rehabilitation exercise quality assessment. Beyond the basic ST-GCN \cite{yan2018spatial}, which includes multiple cascading ST-GCN layers and a global average pooling layer, an 'extended' ST-GCN was developed for exercise-type-specific models. This extended version (1) substitutes global average pooling with LSTM, addressing the loss of subtle features critical for assessing rehabilitation exercise quality, and (2) replaces the fixed adjacency matrix with a dynamically modified one, allowing for adaptive adjustments to the significance of body joints in different exercises. However, these enhancements also considerably increased the models' parameter count and computational complexity.

Zheng et al. \cite{zheng2023skeleton} developed exercise-type-specific assessment models using the vanilla ST-GCN \cite{yan2018spatial} with a reduced number of ST-GCN layers. To make the neural network robust to changes in the positioning of the subject in question and the capture device, a Rotation-Invariant (RI) descriptor, namely the dot product matrix of the human skeleton, was applied to the input to ST-GCN. In addition, to make ST-GCN inferences interpretable and provide visualization of body joints associated with erroneous movements, Gradient-weighted Class Activation Mapping \cite{selvaraju2017grad} was applied to ST-GCN.

Réby et al. \cite{reby2023graph} utilized a combination of ST-GCN with Transformers for developing exercise-type-specific assessment models. The neural network incorporated a spatial self-attention module to understand intra-frame relations between different body joints and a temporal self-attention module for modeling inter-frame interactions. Given the limited number of samples for a specific exercise type in existing rehabilitation exercise quality assessment datasets, the complex network struggled to train effectively and thus did not yield improved results compared with the vanilla ST-GCN \cite{yan2018spatial}.

Mourchid and Salma \cite{mourchid2023mr} proposed a dense spatiotemporal graph convolutional Gated Recurrent Unit (GRU), a combination of ST-GCN, GRU, and Transformer encoder \cite{mourchid2023d} and also an ST-GCN with multiple residual layers and an attention fusion mechanism for exercise-type-specific assessment model development. Li et al. \cite{li2023graph} introduced a graph convolutional Siamese network for the tasks of rehabilitation exercise quality assessment and exercise type classification. The network takes as input a pair consisting of a test exercise and a 'standard' exercise, assessing the correctness of the test exercise in relation to the standard exercise and identifying the exercise type.

Yao et al. \cite{yao2023contrastive} employed a multi-stream adaptive graph convolutional network \cite{shi2020skeleton} and trained it in a contrastive learning setting by minimizing the linear combination of three loss functions. These include a Huber loss function for assessing the difference between the predicted and ground-truth scores, a loss function aimed at reducing the feature distance for samples with similar scores and increasing it for those with significant score differences, and another loss function dedicated to minimizing the deviation in joint attention weights among samples that share similar scores. Despite its complex architecture and training framework, the method performed poorly in rehabilitation exercise quality assessment compared to Deb et al. \cite{deb2022graph}.

The majority of existing methodologies, as outlined above, have involved the development of exercise-type-specific models trained on samples from specific exercise types in a dataset \cite{liao2020review,liao2020deep,deb2022graph,zheng2023skeleton}. In contrast to the current literature, this paper introduces a novel method that leverages training samples encompassing all exercise types in a dataset, resulting in improved rehabilitation exercise quality assessment.

\section{Method}
\label{sec:methodology}
This section details the proposed method for rehabilitation exercise quality assessment. It involves analyzing the sequence of body-joint movements of a subject performing a rehabilitation exercise and outputting a binary value indicating whether the rehabilitation exercise was performed correctly.

\subsection{Background}
\label{sec:background}
\begin{figure}[h]
\centering
\includegraphics[width=\textwidth]{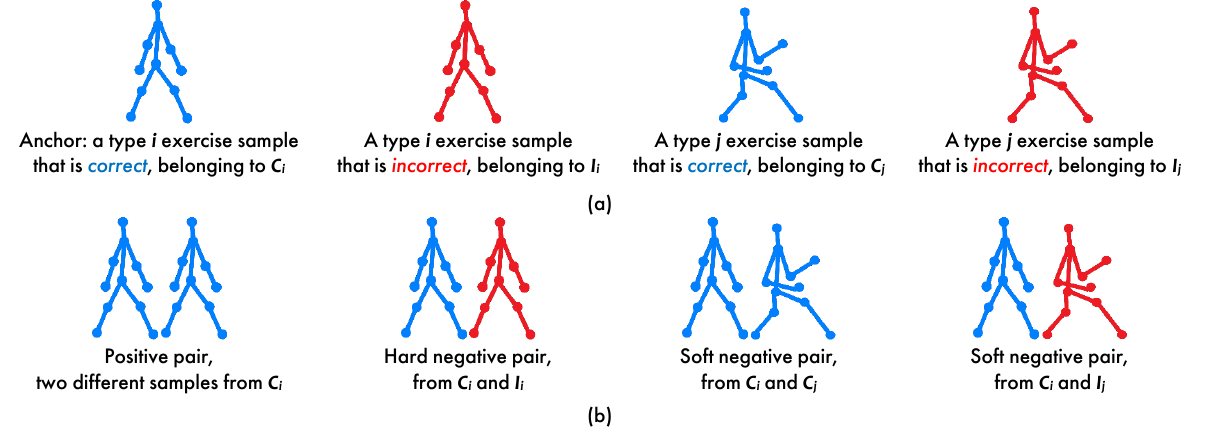}
\caption{(a) Variety of samples in a rehabilitation exercise training mini-batch, featuring different exercise types where each sample may be correct or incorrect, with the leftmost sample designated as the anchor. (b) From left to right, a positive sample pair and its corresponding hard negative sample pair and two soft negative sample pairs.}
\label{fig:fig1}
\end{figure}

Rehabilitation exercise datasets are generally structured as follows. A dataset \( D \) is formed by combining two subsets: \( C \) and \( I \) where \( C \) includes all exercises performed correctly, and \( I \) encompasses those performed incorrectly. Thus, the dataset can be formally defined as \( D = C \cup I \), ensuring that \( C \) and \( I \) are mutually exclusive, indicated by \( C \cap I = \emptyset \). A subset \( D_i \) within \( D \) can be identified as comprising samples of a specific exercise type \( i \). As an extension, \( D_i = C_i \, \cup \, I_i\). Fig. \ref{fig:fig1} (a) illustrates this point.

In existing methodologies \cite{liao2020review,sardari2023artificial,liao2020deep,deb2022graph,zheng2023skeleton,reby2023graph,mourchid2023mr,li2023graph,yao2023contrastive}, the set of all exercise quality assessment models is defined as \( A \), with each model \( A_i \) focused exclusively on the subset \( D_i \). \( A_i \) does not conventionally utilize data outside of its type, \(D \setminus D_i \). This approach is rooted in the fundamental understanding of exercises as either correct or incorrect \cite{capecci2018instrumental,kimore,prmd,irds}. An exercise is deemed correct if the subject successfully completes the prescribed repetitions \cite{capecci2018instrumental,10229850,abedi2023cross}, maintains consistency in execution, adheres to proper technique and movement quality, and ensures proper posture of different body parts \cite{capecci2018instrumental,kimore,liao2020review}. For any particular exercise type, an incorrect exercise is viewed as a suboptimal version of its correct counterpart, signifying either a partial or a complete divergence. However, leveraging the data outside the specific exercise subset, denoted as \( D \setminus D_i \), can be highly beneficial for the assessment model \( A_i \). This benefit arises from the notable difference in the number of samples between \( D_i \) and the entire dataset \( D \), i.e., \( \|D\| \gg \|D_i\| \). Exposing the assessment model \( A_i \) to a broader range of exercise types allows it to better understand and identify the subtleties defining the correct and incorrect execution of exercises.

\subsection{Supervised Contrastive Learning with Hard and Soft Negatives}
\label{method:sup}
Contrastive learning approaches \cite{chen2020simple,supcon} focus on learning representations that effectively differentiate between similar and dissimilar samples. In this context, the aim is to attract similar (positive) sample pairs closer and push away dissimilar (negative) ones within the feature space, thus enhancing the distinction capabilities of the learned representations. We propose a supervised contrastive learning framework with the following categories for sample pairs.

\begin{itemize}
    \item A positive sample pair contains two samples exclusively from \( C_i \).
    \item A hard negative sample pair, as conceptualized in \cite{robinson2021contrastive}, is constructed by coupling samples from \( C_i \) with those from \( I_i \).
    \item A soft negative sample pair is generated by pairing samples of \( C_i \) with those from \(D \setminus D_i\).
\end{itemize}

Sample pairs, as defined above and illustrated in Fig. \ref{fig:fig1} (b), are used to train a neural network consisting of two sub-networks, an encoder \( f(\cdot): {X} \rightarrow \mathbb{R}^{d_f} \) followed by a projection head \( g(\cdot): \mathbb{R}^{d_f} \rightarrow \mathbb{R}^{d_p} \) \cite{chen2020simple,supcon}.

The input to the encoder is a sequence of body joint movements and is defined as \({X} \in \mathbb{R}^{T \times J \times C}\), where \(T\), \(J\), and \(C\) are the sequence length, number of body joints, and the channel size, respectively. When representing each body joint using its horizontal, vertical, and depth coordinates, the channel size is set to 3.

Given a mini-batch of \( N \) tuples \( \{x_{\ell}, y_{\ell}, z_{\ell}\}_{\ell \in [N]} \) where \( x_{\ell} \) denotes the skeleton sequence, \( y_{\ell} \) denotes the exercise type and \( z_{\ell} \) denotes the assessment label as \(z \in \{+,-\}\) for correct and incorrect assessments, respectively. Two independent augmentation functions \( t(\cdot) \) and \( t'(\cdot) \) are applied to the mini-batch to generate a mini-batch of 2-view data samples:

\begin{equation}
\begin{split}
B &= \{(\tilde{x}_{\ell}, y_{\ell}, z_{\ell})\}_{\ell \in [2N]} = \{ ((t(x_{\ell}), y_{\ell}, z_{\ell}), (t'(x_{\ell}), y_{\ell}, z_{\ell})) \}_{\ell \in [N]} .
\end{split}
\label{method:batch}
\end{equation}

Feature embedding is obtained through \(\tilde{\mathbf{v}}_\ell = g(f(\tilde{x}_\ell)), \forall \ell \in [2N]\). Index partitions of \(B\) are formed with \( \beta^+ = \{ \ell \in [2N] \mid z_{\ell} = + \}\), which defines the set of indices where the assessment label \(z_{\ell}\) is correct, and \( \beta_c = \{ \ell \in [2N] \mid y_{\ell} = c \}\), which defines the set of indices where the exercise type \(y_{\ell}\) is \(c\). Furthermore, \( \beta^+_c = \beta^+ \cap \beta_c \) serves as an extension.

Assume that the training sample currently under consideration, known in the context of contrastive learning as the anchor \cite{chen2020simple,supcon}, holds the index \(i\), where \(i \in \beta^+\). In the method being proposed, the anchors are exclusively derived from \(\beta^+\). The contrastive loss is then formulated as follows:



\begin{equation}
\label{method:supcon}
\begin{split}
\mathcal{L} 
&= \sum_{i \in \beta^+}\frac{-1}{\|\beta^+_{y_{i}}\|} \\
& \phantom{=} \sum_{j \in \beta^+_{y_{i}},\, j \neq i}\log\frac{\exp(\mathrm{sim}(\tilde{\mathbf{v}}_{i}, \tilde{\mathbf{v}}_{j}) / \tau)}{\sum_{k \in \beta^-_{y_{i}}}\exp(\mathrm{sim}(\tilde{\mathbf{v}}_{i} ,\tilde{\mathbf{v}}_{k}) / \tau) + \sum_{\ell \neq i}\exp(\mathrm{sim}(\tilde{\mathbf{v}}_{i} ,\tilde{\mathbf{v}}_{\ell}) / \tau)}  
\end{split}
\end{equation}



\noindent
where \( \tau \) is the temperature parameter, and \(\mathrm{sim}(\cdot, \cdot)\) denotes the cosine similarity function between pairs of embedding, as follows:

\begin{equation}
\mathrm{sim}(\tilde{\mathbf{v}}_s, \tilde{\mathbf{v}}_t) = \frac{\tilde{\mathbf{v}}_s \cdot \tilde{\mathbf{v}}_t}{\|\tilde{\mathbf{v}}_s\|_2 \,  \|\tilde{\mathbf{v}}_t\|_2}.
\label{method:cos}
\end{equation}

The numerator in the contrastive loss function in Equation \ref{method:supcon} pertains to positive pairs, which are two correct samples of the same exercise type. The first summation in the denominator pertains to hard negative pairs, consisting of a correct and an incorrect sample of the same exercise type. The second summation in the denominator pertains to soft negative pairs, which include two correct samples of different exercise types and a correct and an incorrect sample of different exercise types. Please refer to Fig. \ref{fig:fig1}.

The contrastive loss function in Equation \ref{method:supcon} is minimized for training the neural network, refer to Fig. 2 (a). The trained network includes the encoder and the projection head, both of which will be employed to generate representations of the input data, refer to Fig. 2 (b) and (c).

\begin{figure}[h]
\centering
\includegraphics[width=1\textwidth]{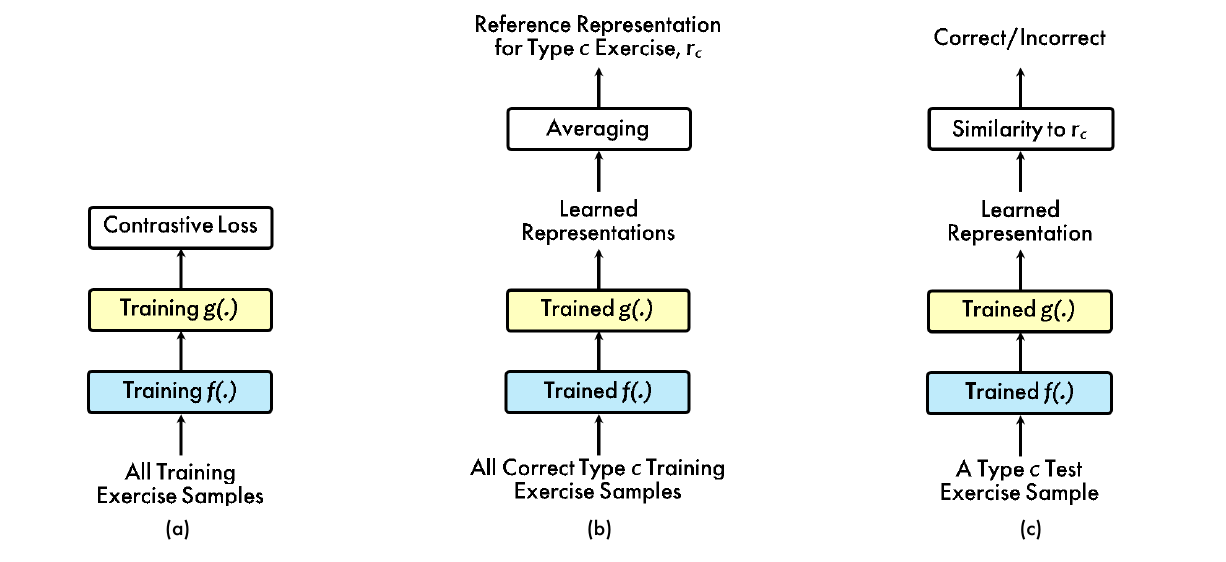}
\caption{(a) Using all training exercise samples with the mini-batches as described in Fig. \ref{fig:fig1} and the supervised contrastive loss function in Equation \ref{method:supcon}, the spatial-temporal graph convolutional network encoder \( f(\cdot) \) and fully-connected projection head \( g(\cdot) \) are trained. (b) Trained \( f(\cdot) \) and \( g(\cdot) \) are used to generate the learned representations for all correct type \( c \) training exercise samples. Weighted averaging of the learned representations results in an exercise-type-specific reference representation for type \( c \) exercise. (c) Inference making by calculating the similarity between the learned representation of a test exercise sample of type \( c \) with the reference representation for type \( c \) exercise.}
\label{fig:fig2}
\end{figure}

\subsection{Inferencing}
\label{method:inference_making}
Rehabilitation exercise quality assessment is conducted based on representations learned through supervised contrastive learning. Drawing inspiration from works in other applications \cite{shehroz,kopuklu2021driver}, the quality of exercise is determined by the degree of similarity between the representations of input data for inference and exercise-specific reference representations, which are derived from correctly performed exercises (described below). Contrary to the traditional contrastive learning approaches \cite{supcon,kopuklu2021driver} where the projection head was discarded post-training, our method retains both the encoder \( f(\cdot)\) and projection head \(g(\cdot)\) during the inference phase \cite{shehroz}.

\(\mathcal{D}\) is defined as the index partition for dataset \(D\), identically to the mini-batch-wise index partitions in subsection \ref{method:sup}. For a given arbitrary exercise type \(c\), a reference representation is generated through a weighted average of the representations of the correct samples for that exercise type. Weight vector \(\mathbf{w}\), is computed according to the inverse variance of each feature dimension from the embedding matrix \(\mathbf{v_{\ell}} = g(f(x_\ell)), \forall \ell \in \mathcal{D}^+_c\),
\begin{equation}
 \mathbf{w} = \frac{1}{\mathrm{Var}(\textbf{V})} \phantom{=} \phantom{=} \phantom{=} \phantom{=} \mathbf{r}_c = \frac{\mathbf{w}}{\ \|\mathbf{w}\|_1} \otimes \sum_{\ell \in \mathcal{D}^+_c} \mathbf{v}_\ell
\label{method:norm}
\end{equation}
where \(\otimes\) denotes Hadamard product. Making inference for a sample with index \(i\) is performed as follows:

\begin{equation}
\hat{\mathbf{p}}_i = \mathrm{sim}(\,g(f(x_i)) \, , \, \mathbf{r}_{y_{i}})
\label{method:eval} 
\end{equation}

\begin{equation}
\hat{\mathbf{z}}_i = 
\begin{cases} 
+ & \text{if } \hat{\mathbf{p}}_i \geq \theta \\
- & \text{otherwise},
\end{cases}
\end{equation}

\noindent
where \(\hat{\mathbf{p}}_i\) denotes the cosine similarity between the representation of the sample indexed by \(i\) and its corresponding reference of the same exercise type, and \(\hat{\mathbf{z}}_i\) represents the binary classification of the sample as correct or incorrect, based on thresholding this similarity with \(\theta\). The threshold is akin to the margin that determines when an exercise is deemed correct. Exercise types with inherently complex criteria for correctness should possess a more relaxed margin for being considered correct.

\section{Experiment}
\label{sec:experiments}
This section evaluates the performance of the proposed method compared to related methods in rehabilitation exercise quality assessment. The results of binary classification and regression for rehabilitation exercise correctness are presented on three publicly available datasets designed for this task.

\subsection{Evaluation Metrics}
\label{evaluation_metrics}
The evaluation metrics for binary classification include accuracy, Area Under the Curve of the Receiver Operating Characteristic curve (AUC ROC), and AUC of the Precision-Recall curve (AUC PR). Furthermore, the representations learned through the proposed method are visualized, illustrating their clustering and corresponding separation. Moreover, the number of parameters in the neural networks of the proposed method is contrasted with those in previously relevant methods. For regression, the evaluation metric employed is Spearman’s rank correlation.

\subsection{Datasets}
\label{datasets}
Experiments were conducted on three publicly available rehabilitation exercise datasets. Each dataset presents unique challenges, further enabling the validation of the proposed learning paradigm.

\begin{itemize}
    \item[\tiny$\bullet$] UI-PRMD \cite{prmd} includes 10 exercise types performed by 10 healthy subjects. Each subject executed 10 repetitions of each exercise, both correctly and incorrectly, on their body's dominant side. Body-joint data were collected using the Kinect sensor at 30 frames per second (fps). The dataset is balanced, featuring a uniform sample distribution across and within exercise types. The 10 types of exercises are deep squat (u01), hurdle step (u02), inline lunge (u03), side lunge (u04), sit to stand (u05), standing active straight leg raise (u06), standing shoulder abduction (u07), standing shoulder extension (u08), standing shoulder internal-external rotation (u09), and standing shoulder scaption (u10).
    \item[\tiny$\bullet$] IRDS \cite{irds} contains 9 exercise types, completed by 15 patients, and 14 healthy subjects. Subjects performed a diverse range of repetitions for each exercise. Exercises have a predetermined side for correct execution. Body joints data was collected with the Kinect One sensor at 30 fps. Some subjects were unable to perform all assigned exercises, resulting in an imbalanced distribution across exercise types. Correct assessments significantly outnumbered incorrect assessments within exercise types. The 9 types of exercises are elbow flexion left (i01), elbow flexion right (i02), shoulder flexion left (i03), shoulder flexion right (i04), shoulder abduction left (i05), shoulder abduction right (i06), shoulder forward elevation (i07), side tap left (i08), and side tap right (i09).
    \item[\tiny$\bullet$] KIMORE \cite{kimore} is a rehabilitation exercise dataset consisting of 5 exercises. It constitutes healthy subjects and patients with motor dysfunctions. Medical professionals assessed all performances with a clinical score for correctness \cite{capecci2018instrumental}, ranging from 0 to 50. Body joints data was collected with the Kinect One sensor at 30 fps. This dataset presents a significant challenge due to its regressive nature, compounded by the limited amount of data available, with only ~70 samples per exercise type across all participants. The 5 types of exercises are lifting of arms (k01), trunk lateral tilt (k02), trunk rotation (k03), pelvis rotation (k04), and squatting (k05).
\end{itemize}

The datasets contain unsegmented and segmented data samples. The segmented set divides each subject's exercise on a repetition basis; that is, samples represent individual repetitions, rather than an entire subject's performance comprising multiple repetitions. Following the literature \cite{zheng2023skeleton}, single-repetition exercises were used for evaluation. All samples were made temporally consistent through down-sampling or up-sampling \cite{zheng2023skeleton}.

\subsection{Experimental Setting}
\label{experimental_setting}
In the datasets for rehabilitation exercise quality assessment \cite{prmd,irds,kimore}, each type of exercise specifies a series of essential body-joint movements to assess the exercise's correctness. These movements are characterized by the joints' trajectory through space and time, enabling the conceptualization of each movement through a sequence of high-order spatial-temporal embeddings. Considering ST-GCN as the optimal model for learning such embeddings \cite{deb2022graph,zheng2023skeleton,reby2023graph,mourchid2023mr,yao2023contrastive}, the encoder's architecture in the proposed method employs an 8-layer ST-GCN as elaborated by Yan et al. \cite{yan2018spatial}. All temporal layers of ST-GCN blocks are downsized using the ResNet bottleneck architecture \cite{actionlet}. A fully-connected layer serves as the projection head, transforming embeddings from a dimensionality of 256 to 128.

The augmentation module comprises spatial, temporal, and spatial-temporal components that preserve the semantic context of data, i.e., maintaining the correctness or incorrectness of the rehabilitation exercise. Drawing inspiration from related works \cite{guo2021contrastive, actionlet, rao2021augmented}, the following augmentations are applied to the training data samples to create two-view training data sample pairs: Spatial shearing and rotation \cite{rao2021augmented}, temporal down- or up-sampling and cropping \cite{rao2021augmented}, and spatial-temporal Gaussian blurring and adding Gaussian noise \cite{rao2021augmented}.

The training was conducted using mini-batches of size 128 for 2000 epochs, with a temperature parameter of 0.1 and a learning rate of 0.001, alongside the ADAM optimizer \cite{kingma2017adam}, using PyTorch \cite{paszke2019pytorch} on a server equipped with 64 GB of RAM and an NVIDIA GeForce RTX 2060 16GB GPU.

\subsection{Experimental Results}
\label{experimental_results}

\subsubsection{Comparison with Previous Works}
\label{comparison_with_previous_works}
As explained in section \ref{sec:related_work_stgcn}, one of the recent pioneering works on rehabilitation exercise quality assessment is by Zheng et al. \cite{zheng2023skeleton}, wherein ST-GCNs were equipped with a RI descriptor to stabilize models against rotational variations in body-joint data. To ensure a fair comparison with relevant works, identical 3:1 training-validation set split as in Zheng et al. \cite{zheng2023skeleton} was adopted. Table \ref{tab:accuracies} depicts the accuracy of the proposed method compared to previous relevant methods on individual exercises and on average on the UI-PRMD \cite{prmd} and IRDS \cite{irds} datasets. In addition to the work by Zheng et al. \cite{zheng2023skeleton}, denoted as ST-GCN with RI in Table \ref{tab:accuracies}, the results of two previous non-ST-GCN methods on body-joint-based action analysis using LSTMs \cite{vanet} and CNNs \cite{pnorm} are also reported \cite{zheng2023skeleton}. The results of the proposed method equipped with the RI descriptor introduced by Zheng et al. \cite{zheng2023skeleton} are also presented in Table \ref{tab:accuracies}. According to Table \ref{tab:accuracies}, on UI-PRMD \cite{prmd}, the proposed method surpassed previous methods in exercises u02, u03, and on average. In nine out of ten exercises, the proposed method achieved an accuracy of 1. On IRDS \cite{irds}, a dataset with imbalanced distributions of samples across correct and incorrect exercises, the proposed method's superiority was more evident, specifically in exercises i01-i04, i06, and i08. The integration of RI \cite{zheng2023skeleton} with the proposed method either mirrored the results of the proposed method or provided a slight boost in accuracy.

\begin{table*}[!ht]
\caption{The accuracy of the proposed method compared to previous methods on u01-u10 in UI-PRMD \cite{prmd} and i01-i09 in IRDS \cite{irds}. Refer to subsection \ref{datasets} for the list of exercises in the datasets. Bolded values denote the best results.}
\label{tab:accuracies}
\resizebox{\textwidth}{!}{
\begin{tabular}{@{}clccccccccccc@{}}
\toprule 
\multicolumn{1}{l}{Dataset}  & \multicolumn{1}{l}{Method} & \multicolumn{1}{l}{u01\textbackslash i01}    & \multicolumn{1}{l}{u02\textbackslash i02} & \multicolumn{1}{l}{u03\textbackslash i03} & \multicolumn{1}{l}{u04\textbackslash i04} & \multicolumn{1}{l}{u05\textbackslash i05} & \multicolumn{1}{l}{u06\textbackslash i06} & \multicolumn{1}{l}{u07\textbackslash i07} & \multicolumn{1}{l}{u08\textbackslash i08} & \multicolumn{1}{l}{u09\textbackslash i09} & \multicolumn{1}{l}{u10} & \multicolumn{1}{l}{average} \\ \midrule

\multirow{5}{*}{UI-PRMD} & \cite{pnorm} & \multicolumn{1}{l}{0.9400} & \multicolumn{1}{l}{0.9600}   & \multicolumn{1}{l}{0.9400}    & \multicolumn{1}{l}{0.9600}    & \multicolumn{1}{l}{0.9800}    & \multicolumn{1}{l}{\textbf{1.0000}}    & \multicolumn{1}{l}{\textbf{1.0000}}    & \multicolumn{1}{l}{0.9800}    & \multicolumn{1}{l}{0.9800}    & \multicolumn{1}{l}{\textbf{1.0000}}    & \multicolumn{1}{l}{0.9740}     \\
                         & \cite{vanet} & \multicolumn{1}{l}{0.9400} & \multicolumn{1}{l}{0.9600}   & \multicolumn{1}{l}{0.9000}    & \multicolumn{1}{l}{0.9800}    & \multicolumn{1}{l}{\textbf{1.0000}}    & \multicolumn{1}{l}{0.9800}    & \multicolumn{1}{l}{\textbf{1.0000}}    & \multicolumn{1}{l}{\textbf{1.0000}}    & \multicolumn{1}{l}{0.9800}    & \multicolumn{1}{l}{\textbf{1.0000}}    & \multicolumn{1}{l}{0.9740}     \\
                         & ST-GCN with RI \cite{zheng2023skeleton} & \multicolumn{1}{l}{\textbf{1.0000}}       & \multicolumn{1}{l}{0.9600}    & \multicolumn{1}{l}{0.9600}    & \multicolumn{1}{l}{\textbf{1.0000}}    & \multicolumn{1}{l}{0.9800}    & \multicolumn{1}{l}{\textbf{1.0000}}    & \multicolumn{1}{l}{\textbf{1.0000}}    & \multicolumn{1}{l}{0.9800}    & \multicolumn{1}{l}{\textbf{1.0000}}    & \multicolumn{1}{l}{\textbf{1.0000}}    & \multicolumn{1}{l}{0.9880}     \\
                         & Proposed  & \multicolumn{1}{l}{\textbf{1.0000}}                      & \multicolumn{1}{l}{\textbf{1.0000}}                   & \multicolumn{1}{l}{0.9800}                  & \multicolumn{1}{l}{\textbf{1.0000}}                   & \multicolumn{1}{l}{0.9800}                  & \multicolumn{1}{l}{0.9800}                  & \multicolumn{1}{l}{\textbf{1.0000}}                   & \multicolumn{1}{l}{\textbf{1.0000}}                   & \multicolumn{1}{l}{\textbf{1.0000}}                   & \multicolumn{1}{l}{\textbf{1.0000}}                   & \multicolumn{1}{l}{0.9940}                   \\
                         & Proposed with RI  & \multicolumn{1}{l}{\textbf{1.0000}}                      & \multicolumn{1}{l}{\textbf{1.0000}}                   & \multicolumn{1}{l}{\textbf{1.0000}}                   & \multicolumn{1}{l}{\textbf{1.0000}}                   & \multicolumn{1}{l}{0.9800}                  & \multicolumn{1}{l}{\textbf{1.0000}}                   & \multicolumn{1}{l}{\textbf{1.0000}}                   & \multicolumn{1}{l}{\textbf{1.0000}}                   & \multicolumn{1}{l}{\textbf{1.0000}}                   & \multicolumn{1}{l}{\textbf{1.0000}}                   & \multicolumn{1}{l}{\textbf{0.9980}}                    \\ \midrule

\multirow{5}{*}{IRDS}    & \cite{pnorm} & \multicolumn{1}{l}{0.9848}                     & \multicolumn{1}{l}{0.9429}                         & \multicolumn{1}{l}{0.9787}                         & \multicolumn{1}{l}{0.9740}                         & \multicolumn{1}{l}{\textbf{1.0000}}                         & \multicolumn{1}{l}{0.9848}                         & \multicolumn{1}{l}{0.9683}                         & \multicolumn{1}{l}{0.9559}                         & \multicolumn{1}{l}{0.9589}                         & \multicolumn{1}{c}{-}                         & \multicolumn{1}{l}{0.9720}                          \\
                         & \cite{vanet} & \multicolumn{1}{l}{0.9697}                            & \multicolumn{1}{l}{0.9571}                         & \multicolumn{1}{l}{0.9681}                         & \multicolumn{1}{l}{0.9740}                         & \multicolumn{1}{l}{0.9714}                         & \multicolumn{1}{l}{0.9848}                         & \multicolumn{1}{l}{\textbf{1.0000}}                         & \multicolumn{1}{l}{0.9412}                         & \multicolumn{1}{l}{0.9452}                          & \multicolumn{1}{c}{-}                         & \multicolumn{1}{l}{0.9680}                          \\
                         & ST-GCN with RI \cite{zheng2023skeleton} & \multicolumn{1}{l}{0.9697}                            & \multicolumn{1}{l}{0.9571}                         & \multicolumn{1}{l}{0.9681}                         & \multicolumn{1}{l}{0.9740}                         & \multicolumn{1}{l}{0.9857}                         & \multicolumn{1}{l}{0.9848}                         & \multicolumn{1}{l}{\textbf{1.0000}}                         & \multicolumn{1}{l}{0.9412}                         & \multicolumn{1}{l}{\textbf{0.9863}}                         & \multicolumn{1}{c}{-} & \multicolumn{1}{l}{0.9741}                          \\
                         & Proposed  & \multicolumn{1}{l}{\textbf{1.0000}}                      & \multicolumn{1}{l}{\textbf{0.9831}}                  & \multicolumn{1}{l}{\textbf{1.0000}}                   & \multicolumn{1}{l}{\textbf{1.0000}}                   & \multicolumn{1}{l}{0.9818}                  & \multicolumn{1}{l}{0.9800}                  & \multicolumn{1}{l}{\textbf{1.0000}}                   & \multicolumn{1}{l}{\textbf{1.0000}}                   & \multicolumn{1}{l}{0.9828}                  & \multicolumn{1}{c}{-} & \multicolumn{1}{l}{\textbf{0.9920}}                   \\
                         & Proposed with RI  & \multicolumn{1}{l}{\textbf{1.0000}}                      & \multicolumn{1}{l}{0.9831}                  & \multicolumn{1}{l}{\textbf{1.0000}}                   & \multicolumn{1}{l}{\textbf{1.0000}}                   & \multicolumn{1}{l}{0.9636}                  & \multicolumn{1}{l}{\textbf{1.0000}}                   & \multicolumn{1}{l}{\textbf{1.0000}}                   & \multicolumn{1}{l}{\textbf{1.0000}}                   & \multicolumn{1}{l}{0.9655}                  & \multicolumn{1}{c}{-}                       & \multicolumn{1}{l}{0.9902}                   \\ \bottomrule
\end{tabular}}
\end{table*}

\begin{table*}[h!]
\caption{AUC ROC of the proposed method on u01-u10 in UI-PRMD \cite{prmd} and i01-i09 in IRDS \cite{irds}. Refer to subsection \ref{datasets} for the list of exercises in the datasets. Bolded values denote the best results.} 
\label{table:roc}
    \resizebox{\textwidth}{!}{
    \begin{tabular}{ccccccccccccc}
    \toprule 
    \multicolumn{1}{l}{Dataset} & Method & u01\textbackslash i01 & u02\textbackslash i02 & u03\textbackslash i03 & u04\textbackslash i04 & u05\textbackslash i05 & u06\textbackslash i06 & u07\textbackslash i07 & u08\textbackslash i08 & u09\textbackslash i09 & u10 & Average \\
    \midrule
    \multirow{2}{*}{UI-PRMD} & 
        Proposed & \textbf{1.0000} & \textbf{1.0000} & 0.9752 & \textbf{1.0000} & \textbf{1.0000} & \textbf{0.9841} & \textbf{1.0000} & \textbf{1.0000} & \textbf{1.0000} & \textbf{1.0000} & 0.9959 \\
        
        & Proposed with RI in \cite{zheng2023skeleton} & \textbf{1.0000} & \textbf{1.0000} & \textbf{1.0000} & \textbf{1.0000} & \textbf{1.0000} & 0.9810 & \textbf{1.0000} & \textbf{1.0000} & \textbf{1.0000} & \textbf{1.0000} & \textbf{0.9981} \\
    \midrule
    \multirow{2}{*}{IRDS} & 
        Proposed & \textbf{1.0000} & \textbf{0.9916} & \textbf{1.0000} & \textbf{1.0000} & 0.9643 & 0.9837 & \textbf{1.0000} & \textbf{1.0000} & \textbf{0.9888} & - & \textbf{0.9920}\\
        
        & Proposed with RI in \cite{zheng2023skeleton} & \textbf{1.0000} & 0.9883 & \textbf{1.0000} & \textbf{1.0000} & \textbf{0.9673} & \textbf{1.0000} & \textbf{1.0000} & \textbf{1.0000} & 0.9636 & - & 0.9910 \\
    \midrule
\end{tabular}
}
\end{table*}

\begin{table*}[h!]
\caption{AUC PR of the proposed method on u01-u10 in UI-PRMD \cite{prmd} and i01-i09 in IRDS \cite{irds}. Refer to subsection \ref{datasets} for the list of exercises in the datasets. Bolded values denote the best results.}
\label{table:pr}
    \resizebox{\textwidth}{!}{
    \begin{tabular}{ccccccccccccc}
    \toprule 
\multicolumn{1}{l}{Dataset} & Method & u01\textbackslash i01 & u02\textbackslash i02 & u03\textbackslash i03 & u04\textbackslash i04 & u05\textbackslash i05 & u06\textbackslash i06 & u07\textbackslash i07 & u08\textbackslash i08 & u09\textbackslash i09 & u10 & Average \\
    \midrule
    \multirow{2}{*}{UI-PRMD} & 
        Proposed & \textbf{1.0000} & \textbf{1.0000} & 0.9827 & \textbf{1.0000} & \textbf{1.0000} & \textbf{0.9873} & \textbf{1.0000} & \textbf{1.0000} & \textbf{1.0000} & \textbf{1.0000} & 0.9970\\
        
        & Proposed with RI in \cite{zheng2023skeleton} & \textbf{1.0000} & \textbf{1.0000} & \textbf{1.0000} & \textbf{1.0000} & \textbf{1.0000} & 0.9840 & \textbf{1.0000} & \textbf{1.0000} & \textbf{1.0000} & \textbf{1.0000} & \textbf{0.9984} \\
    \midrule
    \multirow{2}{*}{IRDS} & 
        Proposed & \textbf{1.0000} & \textbf{0.9976} & \textbf{1.0000} & \textbf{1.0000} & 0.9942 & 0.9963 & \textbf{1.0000} & \textbf{1.0000} & \textbf{0.9986} & - & \textbf{0.9985}\\
        
        & Proposed with RI in \cite{zheng2023skeleton} & \textbf{1.0000} & 0.9965 & \textbf{1.0000} & \textbf{1.0000} & \textbf{0.9952} & \textbf{1.0000} & \textbf{1.0000} & \textbf{1.0000} & 0.9945 & - & 0.9984\\
    \midrule
\end{tabular}
}
\end{table*}

Tables \ref{table:roc}, and \ref{table:pr} respectively display the AUC ROC, and AUC PR of the proposed method on the UI-PRMD \cite{prmd} and IRDS \cite{irds} datasets for individual exercises and on average. The proposed method, with or without RI \cite{zheng2023skeleton}, attained very high AUC ROC and AUC PR values. In particular, the proposed method with RI \cite{zheng2023skeleton} achieved an AUC ROC and AUC PR of 1 for nine out of ten exercises of UI-PRMD \cite{prmd} and for six out of nine exercises of IRDS \cite{irds}, despite the imbalanced distribution of samples between correct and incorrect classes in IRDS \cite{irds}.

\subsubsection{Impact of Retaining the Projection Head}
\label{impact_of_retaining_the_projection_head}
To explore the effectiveness of retaining the projection head during inference, as discussed in subsection \ref{method:inference_making}, the performance of the proposed method equipped with the RI descriptor with and without the projection head on the UI-PRMD dataset \cite{prmd}, was evaluated using five-fold cross-validation and is reported in Table \ref{tab:projection}.

Considering the first two rows in Table \ref{tab:projection}, in all ten exercises of UI-PRMD \cite{prmd}, retaining the projection head for inference led to an accuracy improvement. This improvement was more pronounced in exercises that are uni-lateral or vertically asymmetrical, such as inline lunge (u03) or straight leg raise (u06). In UI-PRMD, the performance of subjects in exercises depends on their dominant side. The single ST-GCN struggles in this scenario because representations for left and right leg raises will be spatially different. This spatial divergence leads to instability in the reference representation, resulting in sub-optimal evaluations. Therefore, projecting side-variant embeddings into an equivalent space is crucial. To reemphasize from the reverse perspective, as shown in Table \ref{tab:projection}, encoder-only representations relatively suffice for symmetrical exercises such as deep squats (u01) and sit-to-stand (u05).

A previous study on driver anomaly detection which kept the projection head within a supervised contrastive learning setting found similar results \cite{shehroz,kopuklu2021driver}. The superior inference results achieved while retaining the projection head can be attributed to the nature of the problem, which involves supervised contrastive learning. The idea of discarding the projection head during inference originated in self-supervised contrastive learning \cite{chen2020simple}, where labels are not available. However, in supervised contrastive learning, where additional information, i.e., labels, is available, the projection head enhances the model's ability to learn more effective representations by adding complexity.

The second and third rows in Table \ref{tab:projection} compare the proposed method with that of Zheng et al. \cite{zheng2023skeleton} using five-fold cross-validation setting. This is different from the comparison in Table \ref{tab:accuracies}, which followed a 3:1 training-validation set split, as in Zheng et al. \cite{zheng2023skeleton}.
The key differences between Zheng et al. \cite{zheng2023skeleton} and the proposed method are twofold: (1) Zheng et al. \cite{zheng2023skeleton} uses 10 distinct ST-GCN-based models, each trained for a specific exercise type. In contrast, the proposed method employs a single ST-GCN-based model that handles all exercise types. (2) The architecture used by Zheng et al. \cite{zheng2023skeleton} consists of an ST-GCN model followed by fully connected layers for classification, without utilizing contrastive learning for model training. On the other hand, the proposed method incorporates contrastive learning with hard and soft negatives. It uses an ST-GCN model as the encoder within the contrastive learning framework, followed by a fully connected network as the projection head. The results show that the proposed contrastive learning-based method, specifically when the projection head is retained, is superior, achieving equal or better accuracy in seven out of ten exercise types.

To further investigate the efficacy of the representations of the encoder and projection head learned through the proposed supervised contrastive learning approach, Support Vector Machines (SVMs) with a Radial Basis Function (RBF) kernel were trained on these representations for exercise quality assessment. For the RBF kernel of the SVMs, the parameters C, and gamma were set to \(1 \), and \( 1/128 = 0.0078 \), respectively. As shown in the last row of Table \ref{tab:projection}, promising results were obtained for U01, U02, and U04, demonstrating the effectiveness of the learned representations through the proposed method; even an SVM applied to these representations can successfully perform exercise quality assessment.


\begin{table*}[h!]
\caption{Exploring the efficacy of learning representations through the proposed supervised contrastive learning approach and the importance of retaining the projection head during inference making. Accuracy of different approaches on u01-u10 in the UI-PRMD \cite{prmd} dataset through five-fold cross-validation is reported. Bolded values denote the best results.}
\label{tab:projection}
\resizebox{\textwidth}{!}{
\begin{tabular}{llcccccccccc}
    \toprule 
    \multicolumn{2}{c}{Accuracy} & u01 & u02 & u03 & u04 & u05 & u06 & u07 & u08 & u09 & u10\\
    \midrule
    \multirow{2}{*}{UI-PRMD}    
    & Proposed (encoder only) & 0.9450 & 0.9050 & 0.8350 & 0.8400 & 0.9450 & 0.8450 & 0.9000 &  0.8850 & 0.8950 & 0.9750 \\
    & Proposed (encoder + projection head) & 0.9900 & 0.9850 & \textbf{0.9850} & \textbf{0.9850} & \textbf{0.9800} & 0.9750 & 0.9800 & \textbf{0.9850} & \textbf{0.9900} & \textbf{0.9950} \\
    & ST-GCN with RI \cite{zheng2023skeleton} & 0.9900 & 0.9900 & 0.9750 & 0.9750 & \textbf{0.9800} & \textbf{0.9900} & \textbf{0.9900} & 0.9800 & \textbf{0.9900} & \textbf{0.9950} \\
    & \begin{tabular}[c]{@{}l@{}} Proposed (encoder + projection head) \\ and SVM as binary classifier \end{tabular} & \textbf{1.0000} & \textbf{1.0000} & 0.9550 & 0.9800 & 0.9450 & 0.9700 & 0.9800 & 0.9700 & 0.9700 & 0.9750 \\
    \midrule
\end{tabular}
}
\end{table*}

\subsubsection{Visualization of Learned Representations}
\label{visualization_of_learned_representations}
\begin{figure}[h]
\centering
\includegraphics[width=1.0\textwidth]{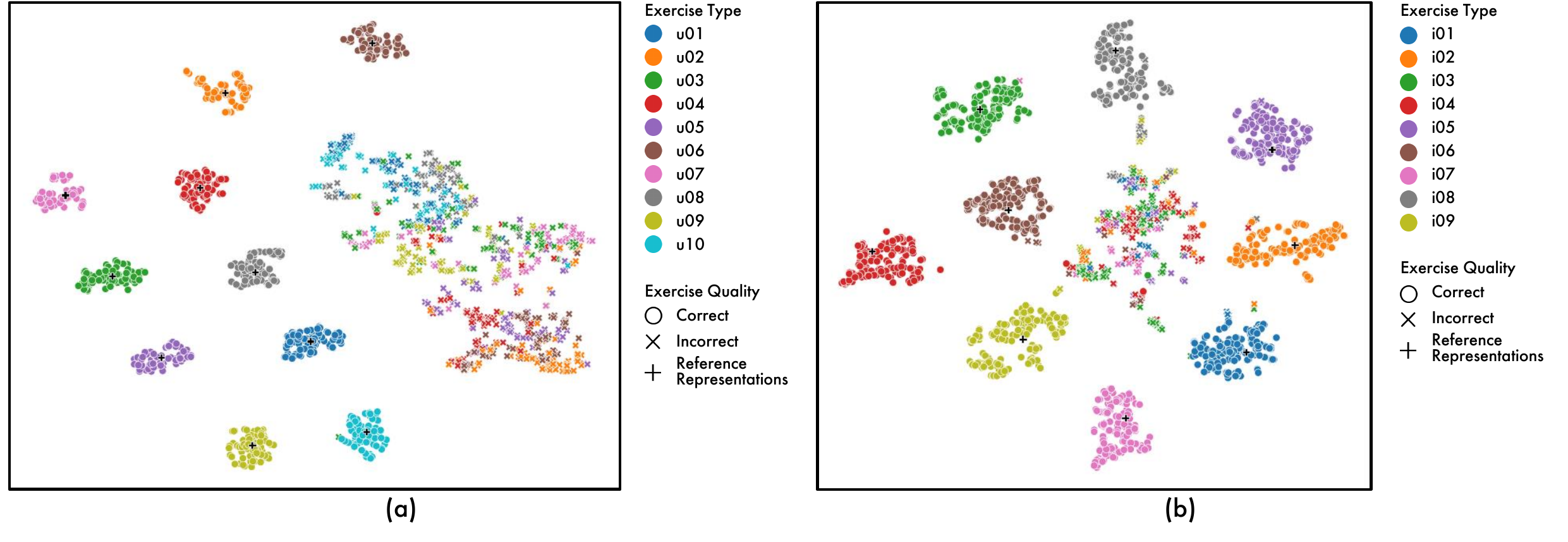}
\caption{t-SNE visualization of representations learned through the proposed supervised contrastive learning approach for (a) UI-PRMD and (b) IRDS datasets. Representations are color-coded on an exercise basis, and ''+'' denotes the reference representations, i.e., cluster centers.}
\label{fig:fig3}
\end{figure}

Applying t-SNE with 2 components and perplexity of 20 \cite{tsne}, the distribution of learned representations in IRDS \cite{irds}, and UI-PRMD \cite{prmd} is illustrated in Fig. \ref{fig:fig3} (a), and (b), respectively. The ST-GCN model, equipped with the RI descriptor and trained through the proposed supervised contrastive learning method, partitions all correct assessments on an exercise basis. Almost all negative assessments are situated outside these clusters, elsewhere in the representation space. The centers of the clusters correspond to the reference representations defined in equation \ref{method:norm}.

\subsubsection{Number of Parameters Compared to Previous Works}
\label{number_of_parameters}
In addition to improving the state-of-the-art in rehabilitation exercise quality assessment datasets \cite{prmd,irds}, a key benefit of the proposed method is its ability to develop a single model for all exercise types within a dataset. This strategy significantly lowers the parameter count of the proposed method in comparison to earlier approaches. For example, the method proposed by Zheng et al. \cite{zheng2023skeleton}, which employs a three-layer ST-GCN network, comprises 818,112 parameters. Given that Zheng et al. \cite{zheng2023skeleton} developed models specific to each exercise type, their model's total parameter count amounts to $818,112 \times 10$ for UI-PRMD \cite{prmd} and $818,112 \times 9$ for IRDS \cite{irds}. The number of parameters in the proposed method, which utilizes an 8-layer ST-GCN network, is 1,249,536. As the proposed method forgoes the training of exercise-type-specific models in favor of a single model trained across the entire dataset, our model is up to 6 times more efficient in terms of parameters, and scales for scenarios that are diverse in exercise types.

\subsubsection{Transfer Learning to KIMORE}
\label{transfer_learning_to_kimore}
The IRDS \cite{irds} and KIMORE \cite{kimore} datasets, both acquired using the Kinect One sensor, feature structurally similar body-joint data; specifically, the adjacency matrices internal to the body joint graph are identical. Therefore, an ST-GCN encoder pre-trained on IRDS \cite{irds} is expected to maintain spatial-temporal relationships when transferred to KIMORE \cite{kimore}. This encoder, alongside the projection head, undergoes training using the proposed method on IRDS \cite{irds}, as outlined in subsection \ref{experimental_setting}.

As explained in subsection \ref{datasets}, the exercise quality assessment problem in IRDS \cite{irds} is a binary classification, categorizing exercises as either correct or incorrect. In contrast, in KIMORE \cite{kimore}, the problem is treated as a regression task, inferring a real-valued number ranging from 0 to 50. To address this inconsistency, while the ST-GCN encoder pre-trained on IRDS \cite{irds} is retained, the projection head pre-trained on IRDS \cite{irds} is replaced with a two-layer fully connected regression network, transforming embeddings from a dimensionality of 256 to 128 and then to a real-valued number. The fully connected regression layer is fine-tuned to individual exercise types in KIMORE \cite{kimore}. In another setting, an untrained ST-GCN encoder is created, a two-layer fully connected regression network with the same structure as the previous setting is added to it, and trained from scratch on individual exercise types in KIMORE \cite{kimore}.

Table \ref{table:spearman} presents Spearman's rank correlations for individual exercise types in KIMORE \cite{kimore} and compares them with Spearman's rank correlations for four previous methods. The correlations are calculated between the models' real-valued outputs and the ground-truth exercise quality annotations in the dataset. As shown in Table \ref{table:spearman}, utilizing transfer learning improves outcomes over training a model from scratch. Except for k02, the approach employing transfer learning outperforms earlier methods for the remaining exercises.

\begin{figure}[h]
\centering
\includegraphics[width=1.0\textwidth]{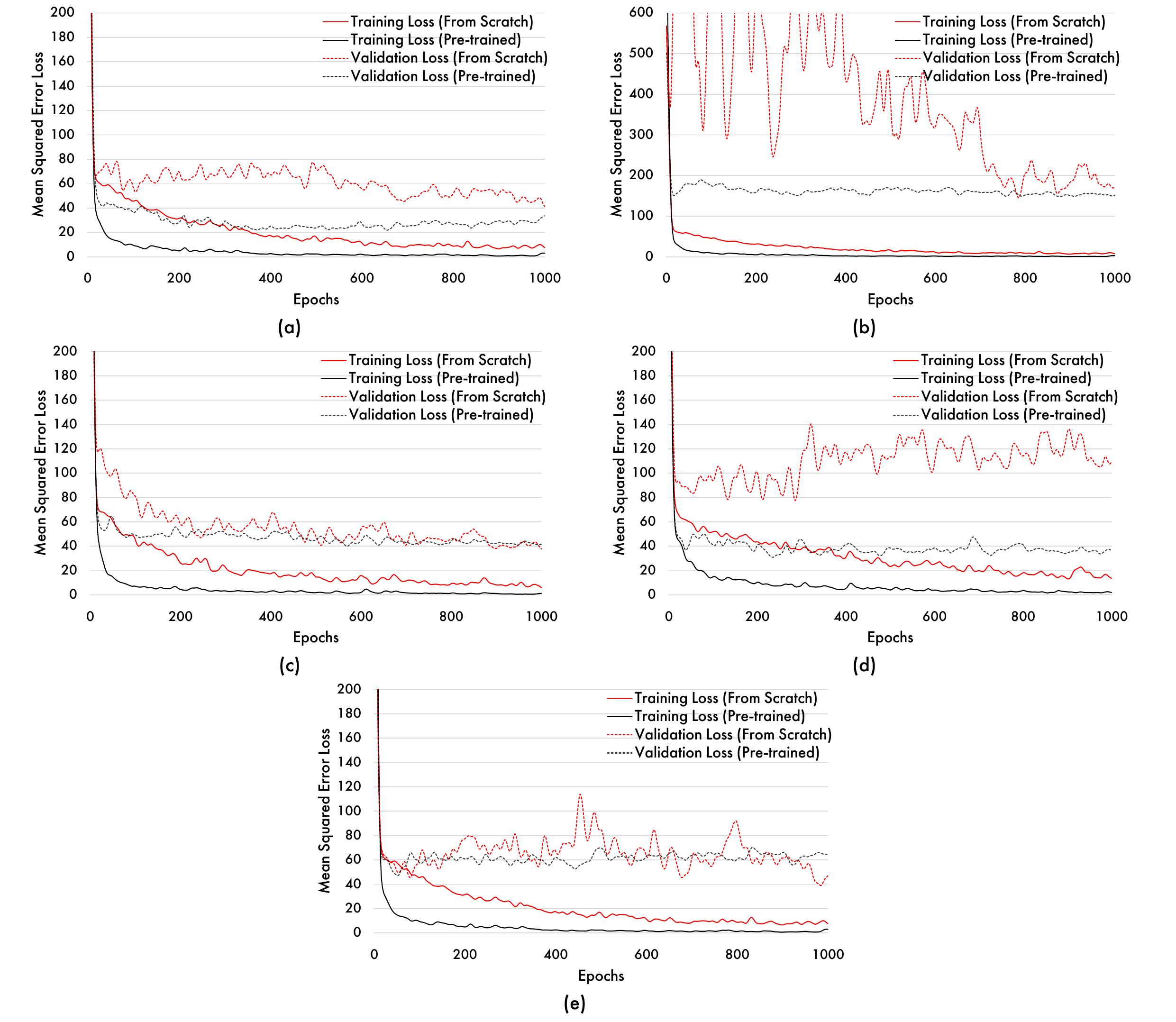}
\caption{Training and validation Mean Squared Error (MSE) loss curves of the proposed method across consecutive epochs for the five exercise types in the KIMORE dataset: (a) lifting of arms, (b) trunk lateral tilt, (c) trunk rotation, (d) pelvis rotation, and (e) squatting. The training was performed in two different settings: training an untrained ST-GCN encoder from scratch, and fine-tuning an ST-GCN encoder pre-trained on the IRDS dataset.}
\label{fig:fig4}
\end{figure}

Fig. \ref{fig:fig4} illustrates the training and validation Mean Squared Error (MSE) loss curves of the proposed method across consecutive epochs for the five exercise types in KIMORE \cite{kimore}. While Fig. \ref{fig:fig4} (a), (b), and (d) demonstrate that the pre-trained ST-GCN achieves improved results, this advantage does not extend to Fig. \ref{fig:fig4} (c) and (e), where the effectiveness of transfer learning is reduced due to exercise misalignment. Certain exercise types in KIMORE \cite{kimore} do not closely align with those in IRDS \cite{irds}. For example, there are no IRDS movements/exercises that encompass squatting, Fig. \ref{fig:fig4} (e). Moreover, IRDS mandates that the torso remain stationary in all exercises, which excludes any trunk rotations, Fig. \ref{fig:fig4} (c). However, the lifting of arms, as shown in Fig. \ref{fig:fig4} (a), may include IRDS movements i03 to i07. Generally, the pre-trained ST-GCN outperforms the un-trained ST-GCN in exercises that share movements with the source dataset. Where there is little or no overlap between the target and source, the performance of the pre-trained ST-GCN should be, at the very least, comparable to that of the un-trained ST-GCN.

\begin{table*}[h!]
\caption{The Spearman’s rank correlation between predictions and ground-truth exercise quality scores in five different exercises, k01-k05, in the KIMORE dataset \cite{kimore} calculated through five-fold cross-validation for two distinct settings of the proposed method compared to the previous methods. Bolded values denote the best results.}
\label{table:spearman}
    \centering
    \resizebox{0.7\textwidth}{!}{
    \centering
    \begin{tabular}{lccccc}
    \toprule 
    Method & k01 & k02 & k03 & k04 & k05 \\
    \midrule
        Capecci et al. \cite{kimore} & 0.44 & 0.41 & 0.46 & 0.62 & 0.30 \\
        Guo and Khan \cite{guo2021exercise} & 0.55 & 0.64 & 0.63 & 0.37 & 0.42 \\
        Karagoz et al. \cite{karagoz2023supervised} & 0.40 & \textbf{0.65} & 0.47 & 0.50 & 0.41 \\
        Abedi et al. \cite{abedi2023cross} & 0.76 & 0.61 & 0.73 & 0.54 & 0.67 \\
        ST-GCN from scratch & 0.72 & 0.57 & 0.77 & 0.74 & 0.72 \\
        ST-GCN fine-tuning (proposed) & \textbf{0.79} & 0.62 & \textbf{0.77} & \textbf{0.80} & \textbf{0.74} \\
    \midrule
\end{tabular}
}
\end{table*}

\section{Conclusion and Future Works}
\label{sec:conclusion}
Our research led to the development of a novel supervised contrastive learning framework for rehabilitation exercise quality assessment. This framework effectively utilizes entire datasets to train a single, versatile model, effectively addressing the challenge of limited samples for individual exercise types in rehabilitation exercise datasets. The successful application of the proposed framework to three publicly available rehabilitation exercise datasets confirms its efficacy and establishes a new standard in accuracy, outperforming existing methods. The proposed model's increased generalizability and reduced parameter count are notable advancements, enhancing efficiency and streamlining integration into practical exercise-based virtual rehabilitation platforms. However, there are some limitations in our work. Similar to previous approaches in this field, it requires a preliminary exercise type classification model before conducting an exercise quality assessment. A significant advancement in the proposed method would be the development of a multitask model capable of simultaneous exercise type classification and assessment, further simplifying the assessment process. Future research directions should include applying this framework across more varied rehabilitation scenarios and refining the model for a greater variety of exercise types. Another major area for improvement is adding interpretability to our method, through techniques such as gradient-based class activation maps for ST-GCNs. This would facilitate an understanding of which body joints and specific timestamps contribute to incorrect exercises. This could then be translated into visual or textual feedback for patients, enhancing the utility and effectiveness of virtual rehabilitation programs.
\\\\
\textbf{Data availability}\\
The datasets analyzed during the current study are publicly available in the following repositories:\\
https://webpages.uidaho.edu/ui-prmd/\\
https://zenodo.org/records/4610859\\
https://vrai.dii.univpm.it/content/kimore-dataset
\\\\
\textbf{Conflict of Interest}\\
The authors declare that they have no conflict of interest.
\\\\
\textbf{Funding}\\
This research was funded by New Frontiers in Research Fund, Canada and TRANSFORM HF Undergraduate Summer Research Program, Canada.
\\\\
\textbf{Author Biography}\\
\textbf{Mark Karlov:} B.Sc. in Electrical and Computer Engineering from University of Toronto. Expert in deep learning for data analysis across various modalities.
\\\\
\textbf{Ali Abedi:} Ph.D. in Electrical Engineering and Computer Science, Postdoctoral Fellow at U of T and UHN. Specializes in machine learning and deep learning technologies.
\\\\
\textbf{Shehroz S. Khan:} Ph.D. in Computer Science, Scientist at UHN, Assistant Professor at U of T. Focuses on AI applications in healthcare and rehabilitation.

\bibliography{sn-bibliography}


\begin{thebibliography}{55}
\ifx \bisbn   \undefined \def \bisbn  #1{ISBN #1}\fi
\ifx \binits  \undefined \def \binits#1{#1}\fi
\ifx \bauthor  \undefined \def \bauthor#1{#1}\fi
\ifx \batitle  \undefined \def \batitle#1{#1}\fi
\ifx \bjtitle  \undefined \def \bjtitle#1{#1}\fi
\ifx \bvolume  \undefined \def \bvolume#1{\textbf{#1}}\fi
\ifx \byear  \undefined \def \byear#1{#1}\fi
\ifx \bissue  \undefined \def \bissue#1{#1}\fi
\ifx \bfpage  \undefined \def \bfpage#1{#1}\fi
\ifx \blpage  \undefined \def \blpage #1{#1}\fi
\ifx \burl  \undefined \def \burl#1{\textsf{#1}}\fi
\ifx \doiurl  \undefined \def \doiurl#1{\url{https://doi.org/#1}}\fi
\ifx \betal  \undefined \def \betal{\textit{et al.}}\fi
\ifx \binstitute  \undefined \def \binstitute#1{#1}\fi
\ifx \binstitutionaled  \undefined \def \binstitutionaled#1{#1}\fi
\ifx \bctitle  \undefined \def \bctitle#1{#1}\fi
\ifx \beditor  \undefined \def \beditor#1{#1}\fi
\ifx \bpublisher  \undefined \def \bpublisher#1{#1}\fi
\ifx \bbtitle  \undefined \def \bbtitle#1{#1}\fi
\ifx \bedition  \undefined \def \bedition#1{#1}\fi
\ifx \bseriesno  \undefined \def \bseriesno#1{#1}\fi
\ifx \blocation  \undefined \def \blocation#1{#1}\fi
\ifx \bsertitle  \undefined \def \bsertitle#1{#1}\fi
\ifx \bsnm \undefined \def \bsnm#1{#1}\fi
\ifx \bsuffix \undefined \def \bsuffix#1{#1}\fi
\ifx \bparticle \undefined \def \bparticle#1{#1}\fi
\ifx \barticle \undefined \def \barticle#1{#1}\fi
\bibcommenthead
\ifx \bconfdate \undefined \def \bconfdate #1{#1}\fi
\ifx \botherref \undefined \def \botherref #1{#1}\fi
\ifx \url \undefined \def \url#1{\textsf{#1}}\fi
\ifx \bchapter \undefined \def \bchapter#1{#1}\fi
\ifx \bbook \undefined \def \bbook#1{#1}\fi
\ifx \bcomment \undefined \def \bcomment#1{#1}\fi
\ifx \oauthor \undefined \def \oauthor#1{#1}\fi
\ifx \citeauthoryear \undefined \def \citeauthoryear#1{#1}\fi
\ifx \endbibitem  \undefined \def \endbibitem {}\fi
\ifx \bconflocation  \undefined \def \bconflocation#1{#1}\fi
\ifx \arxivurl  \undefined \def \arxivurl#1{\textsf{#1}}\fi
\csname PreBibitemsHook\endcsname

\bibitem[\protect\citeauthoryear{{World Health Organization}}{{2023}}]{WHORehabilitation}
\begin{botherref}
\oauthor{\bsnm{{World Health Organization}}}:
{Rehabilitation}.
\url{https://www.who.int/news-room/fact-sheets/detail/rehabilitation}.
{Accessed: January 30, 2023}
({2023})
\end{botherref}
\endbibitem

\bibitem[\protect\citeauthoryear{Dibben et~al.}{2023}]{dibben2023exercise}
\begin{barticle}
\bauthor{\bsnm{Dibben}, \binits{G.O.}},
\bauthor{\bsnm{Faulkner}, \binits{J.}},
\bauthor{\bsnm{Oldridge}, \binits{N.}},
\bauthor{\bsnm{Rees}, \binits{K.}},
\bauthor{\bsnm{Thompson}, \binits{D.R.}},
\bauthor{\bsnm{Zwisler}, \binits{A.-D.}},
\bauthor{\bsnm{Taylor}, \binits{R.S.}}:
\batitle{Exercise-based cardiac rehabilitation for coronary heart disease: a meta-analysis}.
\bjtitle{European heart journal}
\bvolume{44}(\bissue{6}),
\bfpage{452}--\blpage{469}
(\byear{2023})
\end{barticle}
\endbibitem

\bibitem[\protect\citeauthoryear{Frazzitta et~al.}{2013}]{frazzitta2013beneficial}
\begin{barticle}
\bauthor{\bsnm{Frazzitta}, \binits{G.}},
\bauthor{\bsnm{Balbi}, \binits{P.}},
\bauthor{\bsnm{Maestri}, \binits{R.}},
\bauthor{\bsnm{Bertotti}, \binits{G.}},
\bauthor{\bsnm{Boveri}, \binits{N.}},
\bauthor{\bsnm{Pezzoli}, \binits{G.}}:
\batitle{The beneficial role of intensive exercise on parkinson disease progression}.
\bjtitle{American Journal of Physical Medicine and Rehabilitation}
\bvolume{92}(\bissue{6}),
\bfpage{523}--\blpage{532}
(\byear{2013})
\end{barticle}
\endbibitem

\bibitem[\protect\citeauthoryear{Liao et~al.}{2020}]{liao2020review}
\begin{barticle}
\bauthor{\bsnm{Liao}, \binits{Y.}},
\bauthor{\bsnm{Vakanski}, \binits{A.}},
\bauthor{\bsnm{Xian}, \binits{M.}},
\bauthor{\bsnm{Paul}, \binits{D.}},
\bauthor{\bsnm{Baker}, \binits{R.}}:
\batitle{A review of computational approaches for evaluation of rehabilitation exercises}.
\bjtitle{Computers in biology and medicine}
\bvolume{119},
\bfpage{103687}
(\byear{2020})
\end{barticle}
\endbibitem

\bibitem[\protect\citeauthoryear{Shanmugasegaram et~al.}{2012}]{shanmugasegaram2012psychometric}
\begin{barticle}
\bauthor{\bsnm{Shanmugasegaram}, \binits{S.}},
\bauthor{\bsnm{Gagliese}, \binits{L.}},
\bauthor{\bsnm{Oh}, \binits{P.}},
\bauthor{\bsnm{Stewart}, \binits{D.E.}},
\bauthor{\bsnm{Brister}, \binits{S.J.}},
\bauthor{\bsnm{Chan}, \binits{V.}},
\bauthor{\bsnm{Grace}, \binits{S.L.}}:
\batitle{Psychometric validation of the cardiac rehabilitation barriers scale}.
\bjtitle{Clinical rehabilitation}
\bvolume{26}(\bissue{2}),
\bfpage{152}--\blpage{164}
(\byear{2012})
\end{barticle}
\endbibitem

\bibitem[\protect\citeauthoryear{Shirozhan et~al.}{2022}]{shirozhan2022barriers}
\begin{botherref}
\oauthor{\bsnm{Shirozhan}, \binits{S.}},
\oauthor{\bsnm{Arsalani}, \binits{N.}},
\oauthor{\bsnm{Maddah}, \binits{S.S.B.}},
\oauthor{\bsnm{Mohammadi-Shahboulaghi}, \binits{F.}}:
Barriers and facilitators of rehabilitation nursing care for patients with disability in the rehabilitation hospital: A qualitative study.
Frontiers in Public Health
\textbf{10}
(2022)
\end{botherref}
\endbibitem

\bibitem[\protect\citeauthoryear{Combes et~al.}{2018}]{combes2018hospital}
\begin{barticle}
\bauthor{\bsnm{Combes}, \binits{J.-B.}},
\bauthor{\bsnm{Elliott}, \binits{R.F.}},
\bauthor{\bsnm{Sk{\aa}tun}, \binits{D.}}:
\batitle{Hospital staff shortage: the role of the competitiveness of pay of different groups of nursing staff on staff shortage}.
\bjtitle{Applied Economics}
\bvolume{50}(\bissue{60}),
\bfpage{6547}--\blpage{6552}
(\byear{2018})
\end{barticle}
\endbibitem

\bibitem[\protect\citeauthoryear{Ferreira et~al.}{2023}]{ferreira2023usage}
\begin{botherref}
\oauthor{\bsnm{Ferreira}, \binits{R.}},
\oauthor{\bsnm{Santos}, \binits{R.}},
\oauthor{\bsnm{Sousa}, \binits{A.}}:
Usage of auxiliary systems and artificial intelligence in home-based rehabilitation: A review.
Exploring the Convergence of Computer and Medical Science Through Cloud Healthcare,
163--196
(2023)
\end{botherref}
\endbibitem

\bibitem[\protect\citeauthoryear{Krasovsky et~al.}{2020}]{krasovsky2020will}
\begin{barticle}
\bauthor{\bsnm{Krasovsky}, \binits{T.}},
\bauthor{\bsnm{Lubetzky}, \binits{A.V.}},
\bauthor{\bsnm{Archambault}, \binits{P.S.}},
\bauthor{\bsnm{Wright}, \binits{W.G.}}:
\batitle{Will virtual rehabilitation replace clinicians: a contemporary debate about technological versus human obsolescence}.
\bjtitle{Journal of NeuroEngineering and Rehabilitation}
\bvolume{17}(\bissue{1}),
\bfpage{1}--\blpage{8}
(\byear{2020})
\end{barticle}
\endbibitem

\bibitem[\protect\citeauthoryear{Seron et~al.}{2021}]{seron2021effectiveness}
\begin{barticle}
\bauthor{\bsnm{Seron}, \binits{P.}},
\bauthor{\bsnm{Oliveros}, \binits{M.-J.}},
\bauthor{\bsnm{Gutierrez-Arias}, \binits{R.}},
\bauthor{\bsnm{Fuentes-Aspe}, \binits{R.}},
\bauthor{\bsnm{Torres-Castro}, \binits{R.C.}},
\bauthor{\bsnm{Merino-Osorio}, \binits{C.}},
\bauthor{\bsnm{Nahuelhual}, \binits{P.}},
\bauthor{\bsnm{Inostroza}, \binits{J.}},
\bauthor{\bsnm{Jalil}, \binits{Y.}},
\bauthor{\bsnm{Solano}, \binits{R.}}, \betal:
\batitle{Effectiveness of telerehabilitation in physical therapy: a rapid overview}.
\bjtitle{Physical therapy}
\bvolume{101}(\bissue{6}),
\bfpage{053}
(\byear{2021})
\end{barticle}
\endbibitem

\bibitem[\protect\citeauthoryear{Boukhennoufa et~al.}{2022}]{boukhennoufa2022wearable}
\begin{barticle}
\bauthor{\bsnm{Boukhennoufa}, \binits{I.}},
\bauthor{\bsnm{Zhai}, \binits{X.}},
\bauthor{\bsnm{Utti}, \binits{V.}},
\bauthor{\bsnm{Jackson}, \binits{J.}},
\bauthor{\bsnm{McDonald-Maier}, \binits{K.D.}}:
\batitle{Wearable sensors and machine learning in post-stroke rehabilitation assessment: A systematic review}.
\bjtitle{Biomedical Signal Processing and Control}
\bvolume{71},
\bfpage{103197}
(\byear{2022})
\end{barticle}
\endbibitem

\bibitem[\protect\citeauthoryear{Abedi et~al.}{2024}]{abedi2024artificial}
\begin{barticle}
\bauthor{\bsnm{Abedi}, \binits{A.}},
\bauthor{\bsnm{Colella}, \binits{T.J.}},
\bauthor{\bsnm{Pakosh}, \binits{M.}},
\bauthor{\bsnm{Khan}, \binits{S.S.}}:
\batitle{Artificial intelligence-driven virtual rehabilitation for people living in the community: A scoping review}.
\bjtitle{NPJ Digital Medicine}
\bvolume{7}(\bissue{1}),
\bfpage{25}
(\byear{2024})
\end{barticle}
\endbibitem

\bibitem[\protect\citeauthoryear{Sangani et~al.}{2020}]{sangani2020real}
\begin{barticle}
\bauthor{\bsnm{Sangani}, \binits{S.}},
\bauthor{\bsnm{Patterson}, \binits{K.K.}},
\bauthor{\bsnm{Fung}, \binits{J.}},
\bauthor{\bsnm{Lamontagne}, \binits{A.}}, \betal:
\batitle{Real-time avatar-based feedback to enhance the symmetry of spatiotemporal parameters after stroke: Instantaneous effects of different avatar views}.
\bjtitle{IEEE Transactions on Neural Systems and Rehabilitation Engineering}
\bvolume{28}(\bissue{4}),
\bfpage{878}--\blpage{887}
(\byear{2020})
\end{barticle}
\endbibitem

\bibitem[\protect\citeauthoryear{Sardari et~al.}{2023}]{sardari2023artificial}
\begin{botherref}
\oauthor{\bsnm{Sardari}, \binits{S.}},
\oauthor{\bsnm{Sharifzadeh}, \binits{S.}},
\oauthor{\bsnm{Daneshkhah}, \binits{A.}},
\oauthor{\bsnm{Nakisa}, \binits{B.}},
\oauthor{\bsnm{Loke}, \binits{S.W.}},
\oauthor{\bsnm{Palade}, \binits{V.}},
\oauthor{\bsnm{Duncan}, \binits{M.J.}}:
Artificial intelligence for skeleton-based physical rehabilitation action evaluation: A systematic review.
Computers in Biology and Medicine,
106835
(2023)
\end{botherref}
\endbibitem

\bibitem[\protect\citeauthoryear{Fernandez-Cervantes et~al.}{2018}]{fernandez2018virtualgym}
\begin{barticle}
\bauthor{\bsnm{Fernandez-Cervantes}, \binits{V.}},
\bauthor{\bsnm{Neubauer}, \binits{N.}},
\bauthor{\bsnm{Hunter}, \binits{B.}},
\bauthor{\bsnm{Stroulia}, \binits{E.}},
\bauthor{\bsnm{Liu}, \binits{L.}}:
\batitle{Virtualgym: A kinect-based system for seniors exercising at home}.
\bjtitle{Entertainment Computing}
\bvolume{27},
\bfpage{60}--\blpage{72}
(\byear{2018})
\end{barticle}
\endbibitem

\bibitem[\protect\citeauthoryear{Abedi et~al.}{2023a}]{10229850}
\begin{bchapter}
\bauthor{\bsnm{Abedi}, \binits{A.}},
\bauthor{\bsnm{Bisht}, \binits{P.}},
\bauthor{\bsnm{Chatterjee}, \binits{R.}},
\bauthor{\bsnm{Agrawal}, \binits{R.}},
\bauthor{\bsnm{Sharma}, \binits{V.}},
\bauthor{\bsnm{Jayagopi}, \binits{D.}},
\bauthor{\bsnm{Khan}, \binits{S.S.}}:
\bctitle{Rehabilitation exercise repetition segmentation and counting using skeletal body joints}.
In: \bbtitle{2023 20th Conference on Robots and Vision (CRV)},
pp. \bfpage{288}--\blpage{295}.
\bpublisher{IEEE Computer Society},
\blocation{Los Alamitos, CA, USA}
(\byear{2023}).
\doiurl{10.1109/CRV60082.2023.00044} .
\burl{https://doi.ieeecomputersociety.org/10.1109/CRV60082.2023.00044}
\end{bchapter}
\endbibitem

\bibitem[\protect\citeauthoryear{Abedi et~al.}{2023b}]{abedi2023cross}
\begin{botherref}
\oauthor{\bsnm{Abedi}, \binits{A.}},
\oauthor{\bsnm{Malmirian}, \binits{M.}},
\oauthor{\bsnm{Khan}, \binits{S.S.}}:
Cross-modal video to body-joints augmentation for rehabilitation exercise quality assessment.
arXiv preprint arXiv:2306.09546
(2023)
\end{botherref}
\endbibitem

\bibitem[\protect\citeauthoryear{Capecci et~al.}{2019}]{kimore}
\begin{barticle}
\bauthor{\bsnm{Capecci}, \binits{M.}},
\bauthor{\bsnm{Ceravolo}, \binits{M.}},
\bauthor{\bsnm{Ferracuti}, \binits{F.}},
\bauthor{\bsnm{Iarlori}, \binits{S.}},
\bauthor{\bsnm{Monteriu}, \binits{A.}},
\bauthor{\bsnm{Romeo}, \binits{L.}},
\bauthor{\bsnm{Verdini}, \binits{F.}}:
\batitle{The kimore dataset: Kinematic assessment of movement and clinical scores for remote monitoring of physical rehabilitation}.
\bjtitle{IEEE Transactions on Neural Systems and Rehabilitation Engineering}
\bvolume{27}(\bissue{7}),
\bfpage{1436}--\blpage{1448}
(\byear{2019})
\doiurl{10.1109/TNSRE.2019.2923060} .
\bcomment{Epub 2019 Jun 14}
\end{barticle}
\endbibitem

\bibitem[\protect\citeauthoryear{Li et~al.}{2024}]{li2024finerehab}
\begin{bchapter}
\bauthor{\bsnm{Li}, \binits{J.}},
\bauthor{\bsnm{Xue}, \binits{J.}},
\bauthor{\bsnm{Cao}, \binits{R.}},
\bauthor{\bsnm{Du}, \binits{X.}},
\bauthor{\bsnm{Mo}, \binits{S.}},
\bauthor{\bsnm{Ran}, \binits{K.}},
\bauthor{\bsnm{Zhang}, \binits{Z.}}:
\bctitle{Finerehab: A multi-modality and multi-task dataset for rehabilitation analysis}.
In: \bbtitle{Proceedings of the IEEE/CVF Conference on Computer Vision and Pattern Recognition},
pp. \bfpage{3184}--\blpage{3193}
(\byear{2024})
\end{bchapter}
\endbibitem

\bibitem[\protect\citeauthoryear{Capecci et~al.}{2018}]{capecci2018instrumental}
\begin{barticle}
\bauthor{\bsnm{Capecci}, \binits{M.}},
\bauthor{\bsnm{Ceravolo}, \binits{M.G.}},
\bauthor{\bsnm{Ferracuti}, \binits{F.}},
\bauthor{\bsnm{Grugnetti}, \binits{M.}},
\bauthor{\bsnm{Iarlori}, \binits{S.}},
\bauthor{\bsnm{Longhi}, \binits{S.}},
\bauthor{\bsnm{Romeo}, \binits{L.}},
\bauthor{\bsnm{Verdini}, \binits{F.}}:
\batitle{An instrumental approach for monitoring physical exercises in a visual markerless scenario: A proof of concept}.
\bjtitle{Journal of biomechanics}
\bvolume{69},
\bfpage{70}--\blpage{80}
(\byear{2018})
\end{barticle}
\endbibitem

\bibitem[\protect\citeauthoryear{Yan et~al.}{2018}]{yan2018spatial}
\begin{bchapter}
\bauthor{\bsnm{Yan}, \binits{S.}},
\bauthor{\bsnm{Xiong}, \binits{Y.}},
\bauthor{\bsnm{Lin}, \binits{D.}}:
\bctitle{Spatial temporal graph convolutional networks for skeleton-based action recognition}.
In: \bbtitle{Proceedings of the AAAI Conference on Artificial Intelligence},
vol. \bseriesno{32}
(\byear{2018})
\end{bchapter}
\endbibitem

\bibitem[\protect\citeauthoryear{Yao et~al.}{2023}]{yao2023contrastive}
\begin{botherref}
\oauthor{\bsnm{Yao}, \binits{L.}},
\oauthor{\bsnm{Lei}, \binits{Q.}},
\oauthor{\bsnm{Zhang}, \binits{H.}},
\oauthor{\bsnm{Du}, \binits{J.}},
\oauthor{\bsnm{Gao}, \binits{S.}}:
A contrastive learning network for performance metric and assessment of physical rehabilitation exercises.
IEEE Transactions on Neural Systems and Rehabilitation Engineering
(2023)
\end{botherref}
\endbibitem

\bibitem[\protect\citeauthoryear{Zheng et~al.}{2023}]{zheng2023skeleton}
\begin{botherref}
\oauthor{\bsnm{Zheng}, \binits{K.}},
\oauthor{\bsnm{Wu}, \binits{J.}},
\oauthor{\bsnm{Zhang}, \binits{J.}},
\oauthor{\bsnm{Guo}, \binits{C.}}:
A skeleton-based rehabilitation exercise assessment system with rotation invariance.
IEEE Transactions on Neural Systems and Rehabilitation Engineering
(2023)
\end{botherref}
\endbibitem

\bibitem[\protect\citeauthoryear{Deb et~al.}{2022}]{deb2022graph}
\begin{barticle}
\bauthor{\bsnm{Deb}, \binits{S.}},
\bauthor{\bsnm{Islam}, \binits{M.F.}},
\bauthor{\bsnm{Rahman}, \binits{S.}},
\bauthor{\bsnm{Rahman}, \binits{S.}}:
\batitle{Graph convolutional networks for assessment of physical rehabilitation exercises}.
\bjtitle{IEEE Transactions on Neural Systems and Rehabilitation Engineering}
\bvolume{30},
\bfpage{410}--\blpage{419}
(\byear{2022})
\end{barticle}
\endbibitem

\bibitem[\protect\citeauthoryear{Vakanski et~al.}{2018}]{prmd}
\begin{botherref}
\oauthor{\bsnm{Vakanski}, \binits{A.}},
\oauthor{\bsnm{Jun}, \binits{H.-p.}},
\oauthor{\bsnm{Paul}, \binits{D.}},
\oauthor{\bsnm{Baker}, \binits{R.}}:
A data set of human body movements for physical rehabilitation exercises.
Data
\textbf{3}(1)
(2018)
\doiurl{10.3390/data3010002}
\end{botherref}
\endbibitem

\bibitem[\protect\citeauthoryear{Miron et~al.}{2021}]{irds}
\begin{botherref}
\oauthor{\bsnm{Miron}, \binits{A.}},
\oauthor{\bsnm{Sadawi}, \binits{N.}},
\oauthor{\bsnm{Ismail}, \binits{W.}},
\oauthor{\bsnm{Hussain}, \binits{H.}},
\oauthor{\bsnm{Grosan}, \binits{C.}}:
Intellirehabds (irds)—a dataset of physical rehabilitation movements.
Data
\textbf{6}(5)
(2021)
\doiurl{10.3390/data6050046}
\end{botherref}
\endbibitem

\bibitem[\protect\citeauthoryear{Khan et~al.}{2022}]{khan2022inconsistencies}
\begin{botherref}
\oauthor{\bsnm{Khan}, \binits{S.S.}},
\oauthor{\bsnm{Abedi}, \binits{A.}},
\oauthor{\bsnm{Colella}, \binits{T.}}:
Inconsistencies in measuring student engagement in virtual learning--a critical review.
arXiv preprint arXiv:2208.04548
(2022)
\end{botherref}
\endbibitem

\bibitem[\protect\citeauthoryear{Pavllo et~al.}{2019}]{pavllo20193d}
\begin{bchapter}
\bauthor{\bsnm{Pavllo}, \binits{D.}},
\bauthor{\bsnm{Feichtenhofer}, \binits{C.}},
\bauthor{\bsnm{Grangier}, \binits{D.}},
\bauthor{\bsnm{Auli}, \binits{M.}}:
\bctitle{3d human pose estimation in video with temporal convolutions and semi-supervised training}.
In: \bbtitle{Proceedings of the IEEE/CVF Conference on Computer Vision and Pattern Recognition},
pp. \bfpage{7753}--\blpage{7762}
(\byear{2019})
\end{bchapter}
\endbibitem

\bibitem[\protect\citeauthoryear{Lugaresi et~al.}{2019}]{lugaresi2019mediapipe}
\begin{botherref}
\oauthor{\bsnm{Lugaresi}, \binits{C.}},
\oauthor{\bsnm{Tang}, \binits{J.}},
\oauthor{\bsnm{Nash}, \binits{H.}},
\oauthor{\bsnm{McClanahan}, \binits{C.}},
\oauthor{\bsnm{Uboweja}, \binits{E.}},
\oauthor{\bsnm{Hays}, \binits{M.}},
\oauthor{\bsnm{Zhang}, \binits{F.}},
\oauthor{\bsnm{Chang}, \binits{C.-L.}},
\oauthor{\bsnm{Yong}, \binits{M.G.}},
\oauthor{\bsnm{Lee}, \binits{J.}}, et al.:
Mediapipe: A framework for building perception pipelines.
arXiv preprint arXiv:1906.08172
(2019)
\end{botherref}
\endbibitem

\bibitem[\protect\citeauthoryear{Khanghah et~al.}{2023}]{khanghah2023novel}
\begin{barticle}
\bauthor{\bsnm{Khanghah}, \binits{A.B.}},
\bauthor{\bsnm{Fernie}, \binits{G.}},
\bauthor{\bsnm{Fekr}, \binits{A.R.}}:
\batitle{A novel approach to tele-rehabilitation: Implementing a biofeedback system using machine learning algorithms}.
\bjtitle{Machine Learning with Applications}
\bvolume{14},
\bfpage{100499}
(\byear{2023})
\end{barticle}
\endbibitem

\bibitem[\protect\citeauthoryear{Khosla et~al.}{2021}]{supcon}
\begin{botherref}
\oauthor{\bsnm{Khosla}, \binits{P.}},
\oauthor{\bsnm{Teterwak}, \binits{P.}},
\oauthor{\bsnm{Wang}, \binits{C.}},
\oauthor{\bsnm{Sarna}, \binits{A.}},
\oauthor{\bsnm{Tian}, \binits{Y.}},
\oauthor{\bsnm{Isola}, \binits{P.}},
\oauthor{\bsnm{Maschinot}, \binits{A.}},
\oauthor{\bsnm{Liu}, \binits{C.}},
\oauthor{\bsnm{Krishnan}, \binits{D.}}:
Supervised Contrastive Learning
(2021)
\end{botherref}
\endbibitem

\bibitem[\protect\citeauthoryear{Robinson et~al.}{2021}]{robinson2021contrastive}
\begin{botherref}
\oauthor{\bsnm{Robinson}, \binits{J.}},
\oauthor{\bsnm{Chuang}, \binits{C.-Y.}},
\oauthor{\bsnm{Sra}, \binits{S.}},
\oauthor{\bsnm{Jegelka}, \binits{S.}}:
Contrastive Learning with Hard Negative Samples
(2021)
\end{botherref}
\endbibitem

\bibitem[\protect\citeauthoryear{Liao et~al.}{2020}]{liao2020deep}
\begin{barticle}
\bauthor{\bsnm{Liao}, \binits{Y.}},
\bauthor{\bsnm{Vakanski}, \binits{A.}},
\bauthor{\bsnm{Xian}, \binits{M.}}:
\batitle{A deep learning framework for assessing physical rehabilitation exercises}.
\bjtitle{IEEE Transactions on Neural Systems and Rehabilitation Engineering}
\bvolume{28}(\bissue{2}),
\bfpage{468}--\blpage{477}
(\byear{2020})
\end{barticle}
\endbibitem

\bibitem[\protect\citeauthoryear{Bashir et~al.}{2005}]{bashir2005hmm}
\begin{bchapter}
\bauthor{\bsnm{Bashir}, \binits{F.}},
\bauthor{\bsnm{Qu}, \binits{W.}},
\bauthor{\bsnm{Khokhar}, \binits{A.}},
\bauthor{\bsnm{Schonfeld}, \binits{D.}}:
\bctitle{Hmm-based motion recognition system using segmented pca}.
In: \bbtitle{IEEE International Conference on Image Processing 2005},
vol. \bseriesno{3},
p. \bfpage{1288}
(\byear{2005}).
\bcomment{IEEE}
\end{bchapter}
\endbibitem

\bibitem[\protect\citeauthoryear{Lin et~al.}{2023}]{lin2023actionlet}
\begin{bchapter}
\bauthor{\bsnm{Lin}, \binits{L.}},
\bauthor{\bsnm{Zhang}, \binits{J.}},
\bauthor{\bsnm{Liu}, \binits{J.}}:
\bctitle{Actionlet-dependent contrastive learning for unsupervised skeleton-based action recognition}.
In: \bbtitle{Proceedings of the IEEE/CVF Conference on Computer Vision and Pattern Recognition},
pp. \bfpage{2363}--\blpage{2372}
(\byear{2023})
\end{bchapter}
\endbibitem

\bibitem[\protect\citeauthoryear{Guo and Khan}{2021}]{guo2021exercise}
\begin{bchapter}
\bauthor{\bsnm{Guo}, \binits{Q.}},
\bauthor{\bsnm{Khan}, \binits{S.S.}}:
\bctitle{Exercise-specific feature extraction approach for assessing physical rehabilitation}.
In: \bbtitle{4th IJCAI Workshop on AI for Aging, Rehabilitation and Intelligent Assisted Living. IJCAI}
(\byear{2021})
\end{bchapter}
\endbibitem

\bibitem[\protect\citeauthoryear{Karagoz et~al.}{2023}]{karagoz2023supervised}
\begin{barticle}
\bauthor{\bsnm{Karagoz}, \binits{B.}},
\bauthor{\bsnm{Ashraf}, \binits{A.}},
\bauthor{\bsnm{Khan}, \binits{S.}}:
\batitle{Supervised sequential contrastive regression: Improving performance on imbalanced rehabilitation exercises datasets}.
\bjtitle{preprint}
(\byear{2023})
\doiurl{10.13140/RG.2.2.15642.21447}
\end{barticle}
\endbibitem

\bibitem[\protect\citeauthoryear{Zha et~al.}{2024}]{zha2024rank}
\begin{botherref}
\oauthor{\bsnm{Zha}, \binits{K.}},
\oauthor{\bsnm{Cao}, \binits{P.}},
\oauthor{\bsnm{Son}, \binits{J.}},
\oauthor{\bsnm{Yang}, \binits{Y.}},
\oauthor{\bsnm{Katabi}, \binits{D.}}:
Rank-n-contrast: Learning continuous representations for regression.
Advances in Neural Information Processing Systems
\textbf{36}
(2024)
\end{botherref}
\endbibitem

\bibitem[\protect\citeauthoryear{Selvaraju et~al.}{2017}]{selvaraju2017grad}
\begin{bchapter}
\bauthor{\bsnm{Selvaraju}, \binits{R.R.}},
\bauthor{\bsnm{Cogswell}, \binits{M.}},
\bauthor{\bsnm{Das}, \binits{A.}},
\bauthor{\bsnm{Vedantam}, \binits{R.}},
\bauthor{\bsnm{Parikh}, \binits{D.}},
\bauthor{\bsnm{Batra}, \binits{D.}}:
\bctitle{Grad-cam: Visual explanations from deep networks via gradient-based localization}.
In: \bbtitle{Proceedings of the IEEE International Conference on Computer Vision},
pp. \bfpage{618}--\blpage{626}
(\byear{2017})
\end{bchapter}
\endbibitem

\bibitem[\protect\citeauthoryear{R{\'e}by et~al.}{2023}]{reby2023graph}
\begin{bchapter}
\bauthor{\bsnm{R{\'e}by}, \binits{K.}},
\bauthor{\bsnm{Dulau}, \binits{I.}},
\bauthor{\bsnm{Dubrasquet}, \binits{G.}},
\bauthor{\bsnm{Aimar}, \binits{M.B.}}:
\bctitle{Graph transformer for physical rehabilitation evaluation}.
In: \bbtitle{2023 IEEE 17th International Conference on Automatic Face and Gesture Recognition (FG)},
pp. \bfpage{1}--\blpage{8}
(\byear{2023}).
\bcomment{IEEE}
\end{bchapter}
\endbibitem

\bibitem[\protect\citeauthoryear{Mourchid and Slama}{2023a}]{mourchid2023mr}
\begin{bchapter}
\bauthor{\bsnm{Mourchid}, \binits{Y.}},
\bauthor{\bsnm{Slama}, \binits{R.}}:
\bctitle{Mr-stgn: Multi-residual spatio temporal graph network using attention fusion for patient action assessment}.
In: \bbtitle{2023 IEEE 25th International Workshop on Multimedia Signal Processing (MMSP)},
pp. \bfpage{1}--\blpage{6}
(\byear{2023}).
\bcomment{IEEE}
\end{bchapter}
\endbibitem

\bibitem[\protect\citeauthoryear{Mourchid and Slama}{2023b}]{mourchid2023d}
\begin{barticle}
\bauthor{\bsnm{Mourchid}, \binits{Y.}},
\bauthor{\bsnm{Slama}, \binits{R.}}:
\batitle{D-stgcnt: A dense spatio-temporal graph conv-gru network based on transformer for assessment of patient physical rehabilitation}.
\bjtitle{Computers in Biology and Medicine}
\bvolume{165},
\bfpage{107420}
(\byear{2023})
\end{barticle}
\endbibitem

\bibitem[\protect\citeauthoryear{Li et~al.}{2023}]{li2023graph}
\begin{bchapter}
\bauthor{\bsnm{Li}, \binits{C.}},
\bauthor{\bsnm{Ling}, \binits{X.}},
\bauthor{\bsnm{Xia}, \binits{S.}}:
\bctitle{A graph convolutional siamese network for the assessment and recognition of physical rehabilitation exercises}.
In: \bbtitle{International Conference on Artificial Neural Networks},
pp. \bfpage{229}--\blpage{240}
(\byear{2023}).
\bcomment{Springer}
\end{bchapter}
\endbibitem

\bibitem[\protect\citeauthoryear{Shi et~al.}{2020}]{shi2020skeleton}
\begin{barticle}
\bauthor{\bsnm{Shi}, \binits{L.}},
\bauthor{\bsnm{Zhang}, \binits{Y.}},
\bauthor{\bsnm{Cheng}, \binits{J.}},
\bauthor{\bsnm{Lu}, \binits{H.}}:
\batitle{Skeleton-based action recognition with multi-stream adaptive graph convolutional networks}.
\bjtitle{IEEE Transactions on Image Processing}
\bvolume{29},
\bfpage{9532}--\blpage{9545}
(\byear{2020})
\end{barticle}
\endbibitem

\bibitem[\protect\citeauthoryear{Chen et~al.}{2020}]{chen2020simple}
\begin{bchapter}
\bauthor{\bsnm{Chen}, \binits{T.}},
\bauthor{\bsnm{Kornblith}, \binits{S.}},
\bauthor{\bsnm{Norouzi}, \binits{M.}},
\bauthor{\bsnm{Hinton}, \binits{G.}}:
\bctitle{A simple framework for contrastive learning of visual representations}.
In: \bbtitle{International Conference on Machine Learning},
pp. \bfpage{1597}--\blpage{1607}
(\byear{2020}).
\bcomment{PMLR}
\end{bchapter}
\endbibitem

\bibitem[\protect\citeauthoryear{Khan et~al.}{2021}]{shehroz}
\begin{botherref}
\oauthor{\bsnm{Khan}, \binits{S.S.}},
\oauthor{\bsnm{Shen}, \binits{Z.}},
\oauthor{\bsnm{Sun}, \binits{H.}},
\oauthor{\bsnm{Patel}, \binits{A.}},
\oauthor{\bsnm{Abedi}, \binits{A.}}:
Modified supervised contrastive learning for detecting anomalous driving behaviours.
CoRR
\textbf{abs/2109.04021}
(2021)
{\href{https://arxiv.org/abs/2109.04021}{{2109.04021}}}
\end{botherref}
\endbibitem

\bibitem[\protect\citeauthoryear{Kopuklu et~al.}{2021}]{kopuklu2021driver}
\begin{bchapter}
\bauthor{\bsnm{Kopuklu}, \binits{O.}},
\bauthor{\bsnm{Zheng}, \binits{J.}},
\bauthor{\bsnm{Xu}, \binits{H.}},
\bauthor{\bsnm{Rigoll}, \binits{G.}}:
\bctitle{Driver anomaly detection: A dataset and contrastive learning approach}.
In: \bbtitle{Proceedings of the IEEE/CVF Winter Conference on Applications of Computer Vision},
pp. \bfpage{91}--\blpage{100}
(\byear{2021})
\end{bchapter}
\endbibitem

\bibitem[\protect\citeauthoryear{Lin et~al.}{2023}]{actionlet}
\begin{botherref}
\oauthor{\bsnm{Lin}, \binits{L.}},
\oauthor{\bsnm{Zhang}, \binits{J.}},
\oauthor{\bsnm{Liu}, \binits{J.}}:
Actionlet-Dependent Contrastive Learning for Unsupervised Skeleton-Based Action Recognition
(2023)
\end{botherref}
\endbibitem

\bibitem[\protect\citeauthoryear{Guo et~al.}{2021}]{guo2021contrastive}
\begin{botherref}
\oauthor{\bsnm{Guo}, \binits{T.}},
\oauthor{\bsnm{Liu}, \binits{H.}},
\oauthor{\bsnm{Chen}, \binits{Z.}},
\oauthor{\bsnm{Liu}, \binits{M.}},
\oauthor{\bsnm{Wang}, \binits{T.}},
\oauthor{\bsnm{Ding}, \binits{R.}}:
Contrastive Learning from Extremely Augmented Skeleton Sequences for Self-supervised Action Recognition
(2021)
\end{botherref}
\endbibitem

\bibitem[\protect\citeauthoryear{Rao et~al.}{2021}]{rao2021augmented}
\begin{botherref}
\oauthor{\bsnm{Rao}, \binits{H.}},
\oauthor{\bsnm{Xu}, \binits{S.}},
\oauthor{\bsnm{Hu}, \binits{X.}},
\oauthor{\bsnm{Cheng}, \binits{J.}},
\oauthor{\bsnm{Hu}, \binits{B.}}:
Augmented Skeleton Based Contrastive Action Learning with Momentum LSTM for Unsupervised Action Recognition
(2021)
\end{botherref}
\endbibitem

\bibitem[\protect\citeauthoryear{Kingma and Ba}{2017}]{kingma2017adam}
\begin{botherref}
\oauthor{\bsnm{Kingma}, \binits{D.P.}},
\oauthor{\bsnm{Ba}, \binits{J.}}:
Adam: A Method for Stochastic Optimization
(2017)
\end{botherref}
\endbibitem

\bibitem[\protect\citeauthoryear{Paszke et~al.}{2019}]{paszke2019pytorch}
\begin{botherref}
\oauthor{\bsnm{Paszke}, \binits{A.}},
\oauthor{\bsnm{Gross}, \binits{S.}},
\oauthor{\bsnm{Massa}, \binits{F.}},
\oauthor{\bsnm{Lerer}, \binits{A.}},
\oauthor{\bsnm{Bradbury}, \binits{J.}},
\oauthor{\bsnm{Chanan}, \binits{G.}},
\oauthor{\bsnm{Killeen}, \binits{T.}},
\oauthor{\bsnm{Lin}, \binits{Z.}},
\oauthor{\bsnm{Gimelshein}, \binits{N.}},
\oauthor{\bsnm{Antiga}, \binits{L.}},
\oauthor{\bsnm{Desmaison}, \binits{A.}},
\oauthor{\bsnm{K{\"{o}}pf}, \binits{A.}},
\oauthor{\bsnm{Yang}, \binits{E.Z.}},
\oauthor{\bsnm{DeVito}, \binits{Z.}},
\oauthor{\bsnm{Raison}, \binits{M.}},
\oauthor{\bsnm{Tejani}, \binits{A.}},
\oauthor{\bsnm{Chilamkurthy}, \binits{S.}},
\oauthor{\bsnm{Steiner}, \binits{B.}},
\oauthor{\bsnm{Fang}, \binits{L.}},
\oauthor{\bsnm{Bai}, \binits{J.}},
\oauthor{\bsnm{Chintala}, \binits{S.}}:
Pytorch: An imperative style, high-performance deep learning library.
CoRR
\textbf{abs/1912.01703}
(2019)
{\href{https://arxiv.org/abs/1912.01703}{{1912.01703}}}
\end{botherref}
\endbibitem

\bibitem[\protect\citeauthoryear{Zhang et~al.}{2019}]{vanet}
\begin{botherref}
\oauthor{\bsnm{Zhang}, \binits{P.}},
\oauthor{\bsnm{Lan}, \binits{C.}},
\oauthor{\bsnm{Xing}, \binits{J.}},
\oauthor{\bsnm{Zeng}, \binits{W.}},
\oauthor{\bsnm{Xue}, \binits{J.}},
\oauthor{\bsnm{Zheng}, \binits{N.}}:
View Adaptive Neural Networks for High Performance Skeleton-based Human Action Recognition
(2019)
\end{botherref}
\endbibitem

\bibitem[\protect\citeauthoryear{Tasnim et~al.}{2020}]{pnorm}
\begin{botherref}
\oauthor{\bsnm{Tasnim}, \binits{N.}},
\oauthor{\bsnm{Islam}, \binits{M.M.}},
\oauthor{\bsnm{Baek}, \binits{J.-H.}}:
Deep learning-based action recognition using 3d skeleton joints information.
Inventions
\textbf{5}(3)
(2020)
\doiurl{10.3390/inventions5030049}
\end{botherref}
\endbibitem

\bibitem[\protect\citeauthoryear{van~der Maaten and Hinton}{2008}]{tsne}
\begin{barticle}
\bauthor{\bsnm{Maaten}, \binits{L.}},
\bauthor{\bsnm{Hinton}, \binits{G.}}:
\batitle{Visualizing data using t-sne}.
\bjtitle{Journal of Machine Learning Research}
\bvolume{9}(\bissue{86}),
\bfpage{2579}--\blpage{2605}
(\byear{2008})
\end{barticle}
\endbibitem

\end{thebibliography}

\end{document}